%% file: main.tex
\documentclass[10pt,twocolumn,letterpaper]{article}

\usepackage[pagenumbers]{cvpr} % To force page numbers, e.g. for an arXiv version

\definecolor{cvprblue}{rgb}{0.21,0.49,0.74}
\usepackage[pagebackref,breaklinks,colorlinks,allcolors=cvprblue]{hyperref}
\usepackage{amsmath,amssymb} % 常用数学扩展
\usepackage{bm}  
\usepackage{indentfirst}  %缩进
\usepackage{graphicx}
\usepackage{caption}
\usepackage{lipsum}
\usepackage{upgreek}   % 正体希腊字母
\usepackage{enumitem}
\usepackage{booktabs}
\usepackage{multirow}
\usepackage{graphicx}
\usepackage{xcolor}
\usepackage{array}
\usepackage{colortbl}
\newcommand{\best}[1]{\textbf{#1}}
\def\paperID{10942} % *** Enter the Paper ID here
\def\confName{CVPR}
\def\confYear{2026}

\title{TPCNet: Triple physical constraints for Low-light Image Enhancement}

\author{
  Jing-Yi Shi$^{1,2}$ \quad
  Ming-Fei Li$^{1,2}$\thanks{Corresponding author. Email: mf\_li@iphy.ac.cn} \quad
  Ling-An Wu$^{1,2}$ \\
  $^{1}$Institute of Physics, China Academy of Sciences, Beijing 100190, China \\
  $^{2}$School of Physical Sciences, University of Chinese Academy of Sciences, Beijing 100049, China \\
}

\begin{document}
\setlength{\abovecaptionskip}{2mm} %表格到caption的距离
\setlength{\belowcaptionskip}{-5mm}   % caption 与正文的距离
% \setlength{\textfloatsep}{-6pt plus 1pt minus 2pt} % 图与正文间距
% \setlength{\floatsep}{6pt plus 1pt minus 2pt}     % 图与图之间的间距
% \setlength{\intextsep}{6pt plus 1pt minus 2pt}    % 文中插图的间距
% \setlength{\abovecaptionskip}{3pt}                % 图与标题之间
% \setlength{\belowcaptionskip}{0pt}                % 标题与正文之间
% \twocolumn[{%
% \renewcommand\twocolumn[1][]{#1}%
\maketitle
\input{sec/0_abstract}    
\input{sec/1_intro}
\input{sec/2_relate}
\input{sec/3_method}

{
    \small
    \bibliographystyle{ieeenat_fullname}
    \bibliography{main}
}


\end{document}

%% file: sec/0_abstract.tex
\begin{abstract}
Low-light image enhancement is an essential computer vision task to improve image contrast and to decrease the effects of color bias and noise. Many existing interpretable deep-learning algorithms exploit the Retinex theory as the basis of model design. However, previous Retinex-based algorithms, that consider reflected objects as ideal Lambertian ignore  specular reflection in the modeling process and construct the physical constraints in image space, limiting generalization of the model. To address this issue, we preserve the specular reflection coefficient and reformulate the original physical constraints in the imaging process based on the Kubelka-Munk theory, thereby constructing constraint relationship between illumination, reflection, and detection, the so-called triple physical constraints (TPCs) theory. Based on this theory, the physical constraints are constructed in the feature space of the model to obtain the TPC network (TPCNet). Comprehensive quantitative and qualitative benchmark and ablation experiments confirm that these constraints effectively improve the performance metrics and visual quality without introducing new parameters, and demonstrate that our TPCNet outperforms other state-of-the-art methods on 10 datasets. Code is available at \url{https://github.com/2020shijingyi/TPCNet}
\end{abstract}

%% file: sec/1_intro.tex
\begin{figure}[t]
  \centering
   \includegraphics[width=\linewidth]{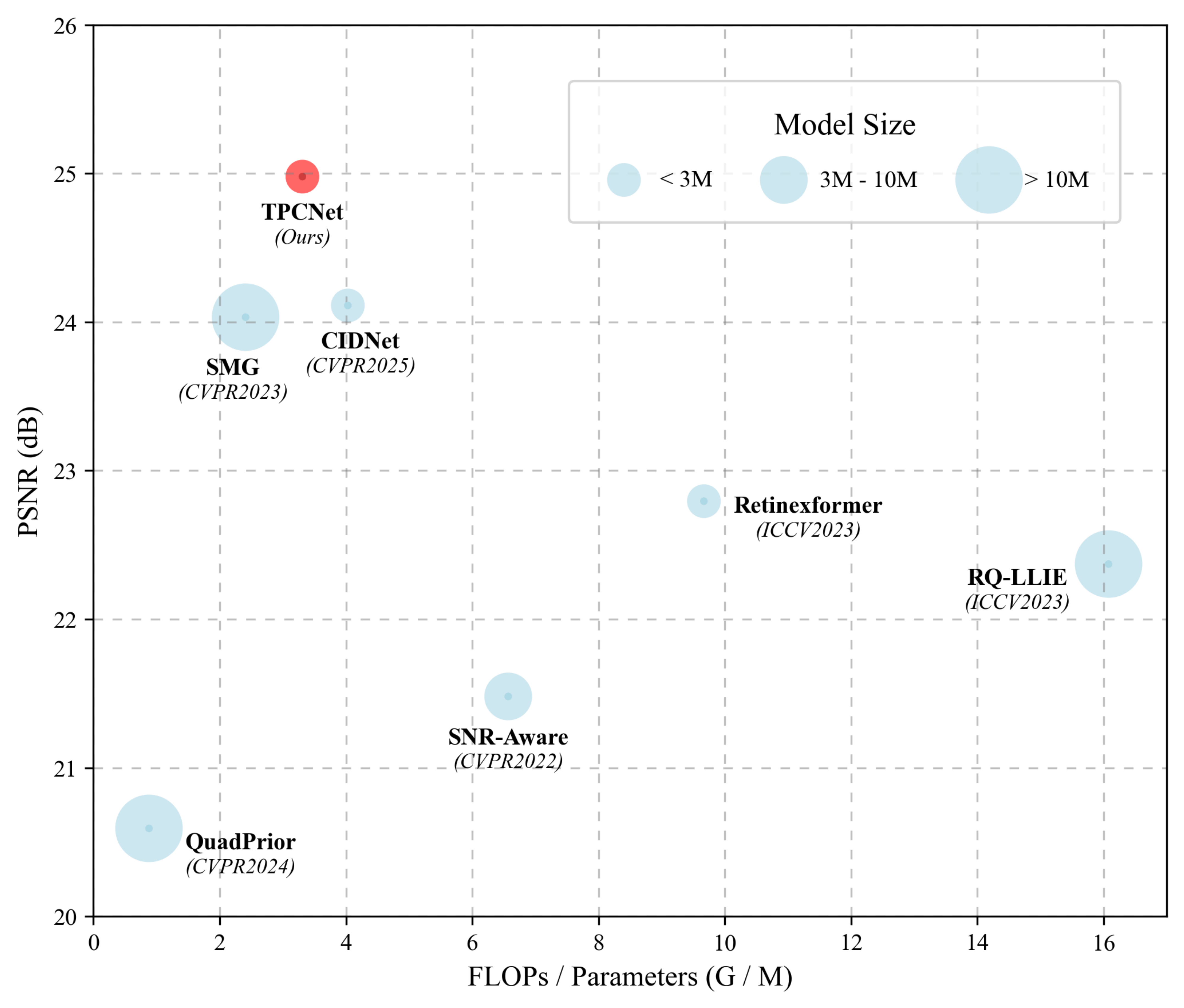}
   \caption{Our TPCNet shows its efficient design and outstanding performance compared to recent SOTA methods, including CIDNet (HVI \cite{6}, CVPR2025), QuadPrior (Quadruple Priors \cite{2}, CVPR2024), SMG (SMG-LLIE \cite{36}, CVPR2023 ), Retinexformer \cite{12} (ICCV2023), RQ-LLIE \cite{37} (ICCV2023), and SNR-Aware (SNR-Net \cite{16}, CVPR2022) for Low-light Image enhancement on LOLv2-real \cite{29} benchmarks.}
   \label{fig:1}
\end{figure}
\section{Introduction}
\label{sec:intro}
A major challenge for computer vision is how to overcome the low signal-to-noise ratio and inaccurate colors of images caused by underexposure \cite{17,18}. For this reason, low-light image enhancement (LLIE), which aims to improve image contrast while minimizing the effects of color bias, is an essential task \cite{11_39}. During the past decades, conventional non-learning methods, like gamma correction \cite{19} and histogram equalization \cite{20,21}, have been widely applied for low-light enhancement, performing transformation on pixel values or adjusting their distribution to realize dynamic range expansion and contrast adjustments. In addition, various Retinex-based methods \cite{11_39,22} decompose low-light images into illuminated and reflected components and enhance the reflected component or optimize the illuminated features to realize enhancement. The above methods all exhibit interpretable  physical processes, but these artificially designed constraints or priors may cause color bias and noise as their adaptability is insufficient for LLIE when dealing with complex illumination.

With the development of deep learning, convolutional neural networks (CNNs) and vision transformers (ViTs), have been applied in various tasks and have made significant advancements \cite{23,24,25,26,27}. For LLIE, deep learning combined with an interpretable physical model has gradually evolved as an effective approach, rather than viewing the learnable modules as a simple black-box. Recently, COMO-ViT \cite{28} constructed an illumination-aware gamma correction module whose gamma factor is calculated by the learnable modules to avoid the bias caused by manual design. Meanwhile, the deep learning framework inspired by the Retinex theory utilizes different CNNs (or ViTs) as the independent estimators for illumination $\mathbf{\hat L}$ and reflectivity $\mathbf{\hat R}$ and computes the enhanced images by $\mathbf{\hat L} \odot \mathbf{\hat R}$ ($\odot$ denotes the element-wise multiplication). Under the guidance of this theory, various methods \cite{3,8,10,12} have been proposed; however, Retinex-based learnable methods typically regard the image as the invariant and decompose it into reflectivity and illumination components. To construct the explicit physical constraint ($\mathbf{I} = \mathbf{\hat L} \odot \mathbf{\hat R}$) between reflectivity and illumination, which is based on actual space ($\mathbf{R} \in {{\cal R}^{3 \times H \times W}}$ and $\mathbf{L} \in {{\cal R}^{1 \times H \times W}}$) and relies on input $\mathbf{I}$, it is necessary to supervise reflectivity (or illumination) features during training with a complex loss function \cite{9,25}. 

Inspired by physical quadruple priors \cite{1_7,2}, which extract illumination invariant features from input, we exploit the intrinsic features in the imaging process to construct implicit constraints, without complex loss functions for supervising the process variables during training, and so can decrease the difficulty in designing the loss functions. Specifically, we reformulate the physical constraints in the imaging process based on the Kubelka-Munk (KM) theory \cite{32,33} and construct the triple physical constraint (TPC) among the detection $\mathbf{\hat E}$ (of the reflected light), illumination $\mathbf{\hat e}$ (by the light source), and reflectivity $\mathbf{\hat R}$ (of the material). Different from the previous Retinex theory, which only considers ideal Lambertian reflection, we retain the specular reflection coefficient and replace it with learnable variables, thus allowing the $\mathbf{\hat e}$ to characterize two complementary illumination components $\mathbf{\hat L}$ and $ \mathbf{\hat{\bar L}}$. Guided by this TPC theory, we design the reflectivity feature estimators (RFE) and light feature estimators (LFE) to estimate the initial $\mathbf{\hat R}$ and $\mathbf{\hat L}$, respectively, while the initial estimates are enhanced by utilizing the backbone structure, a Dual-Stream Cross-Guided Transformer (DCGT) whose core component is a Cross-Guided Multi-Head Self-Attention(CG-MSA). Based on the relationship between $\mathbf{\hat E}$, $\mathbf{\hat e}$, and $\mathbf{\hat R}$, we construct constraints in an implicit feature space (${{\cal R}^{C \times H \times W}}$) rather than in the image space, which can be easily extended to different datasets and improves model robustness. Finally, we derive our method, TPCNet, by combining the core components RFE, LFE, and DCGT with the Color-Association Mechanism (CAM), which is exploited to decrease color bias. As shown in Fig. \ref{fig:1}, our TPCNet outperforms recent state-of-the-art (SOTA) lightweight approaches on the real-image dataset LOL-v2-real \cite{29}.

%------------------------------------------------------------------------

Our contributions can be summarized as follows:
\setlength{\parskip}{0.3em}  % 稍微小一点
\begin{itemize}[noitemsep, topsep=-5pt, leftmargin=*]
    \item We reformulate the origin of physical constraints in the imaging process, based on the KM theory, and construct the TPC among the detection, illumination, and reflectivity. This offers an effective and interpretable direction to design robust networks.
    \item We further propose the TPCNet, a novel LLIE CNN-Transformer framework, which establishes the physical constraints in an implicit feature space to improve its robustness and generalization. Although the TPCNet possesses not so many parameters (2.62M), it exhibits computationally-efficient capacity (only 8.68GFLOPs), benefiting from the efficient design of CG-MSA, which consumes only 25\% of the computation of conventional multi-head self-attention (MSA) with the same input size.
    \item Quantitative and qualitative experiments results demonstrate that our TPCNet outperforms other SOTA approaches on different metrics across 10 datasets. 
\end{itemize}

%% file: sec/2_relate.tex
\section{Related Work}
\label{sec:relate}
\subsection{Low-Light Image Enhancement via Physical Modelling}
\textbf{Retinex-based model.} The Retinex theory \cite{4}, a human visual simulation model widely applied in LLIE, is an effective enhancement algorithm. It regards the observed image as one that can be decomposed into illuminated and reflected components: $\mathbf{I} = \mathbf{\hat L} \odot \mathbf{\hat R}$, where $\mathbf{\hat L}$  and $\mathbf{\hat R}$ refer to the illumination and the reflectance, respectively. Early approaches based on Retinex viewed the reflectance as the final enhanced result, usually causing the result to look unnatural and to be excessively enhanced, \eg, single-scale Retinex \cite{22}. Later, Guo \etal proposed the LIME algorithm \cite{11_39}, which enhances the image via illumination map estimation, improving the visual aesthetics and imaging efficiency \cite{22,30}. Recently, deep learning combined with interpretable physical modeling has been increasingly seen as an effective direction for LLIE. RetinexNet \cite{3}, which combines the Retinex model with data-driven schemes, realizes more accurate reconstruction and more efficient image decomposition than conventional Retinex algorithms by utilizing elaborately hand-crafted loss functions. The above Retinex Imaging model assumes images are corruption-free for decomposing. Corruption-free images cannot be achieved in the inadequate lighting scenes and flaws in the imaging system. To decompose the images the Retinex model assumes that they are corruption-free, but such images cannot be achieved when the lighting is inadequate and there are flaws in the imaging system. To overcome this limitation, Li \etal. \cite{31} proposed a robust Retinex model and pointed out that for real environments the model should include a noise term $\mathbf{N}$ as follows: $\mathbf{I} = \mathbf{\hat L} \odot \mathbf{\hat R} + \mathbf{N}$. Then, Cai \etal. \cite{12} formulated a simple yet one-stage Retinex-based framework $\mathbf{I} = (\mathbf{\hat L} + \mathbf{{\delta _{\hat L}}}) \odot (\mathbf{\hat R} +\mathbf{{\delta _{\hat R}}})$ which introduced perturbation terms $\mathbf{{\delta _{\hat L}}}$ and $\mathbf{{\delta _{\hat R}}}$ for illumination and reflectance, respectively. However, these frameworks construct physical constraints in real space and rely on input features, so the constraints formed by learning generalize poorly to other datasets.

\textbf{Kubelka–Munk-based model}. The KM theory \cite{32}, which assumes that light within a material is isotropically scattered and characterizes the material layers by a wavelength-dependent scatter coefficient and absorption coefficient, is used for materials that both reflect and transmit light. As a general image formation model, it is adopted for studying the reflectance of light in real-world scenes, based on which Geusebroek \etal. proposed a photometric reflectance model to formulate color invariance \cite{33}. Recently, Wang \etal\cite{2} devised illumination invariance, a physical quadruple prior, which originated from the KM theory, and can realize robust zero-reference LLIE. However, the KM model is mainly applied to calculate the invariances of the imaging process, while Retinex is a special case of this theory. The potential of the KM theory for LLIE to realize image decomposition or construct physical constraints has not yet been fully explored.

\begin{figure*}[t]
  \centering
   \includegraphics[width=\linewidth]{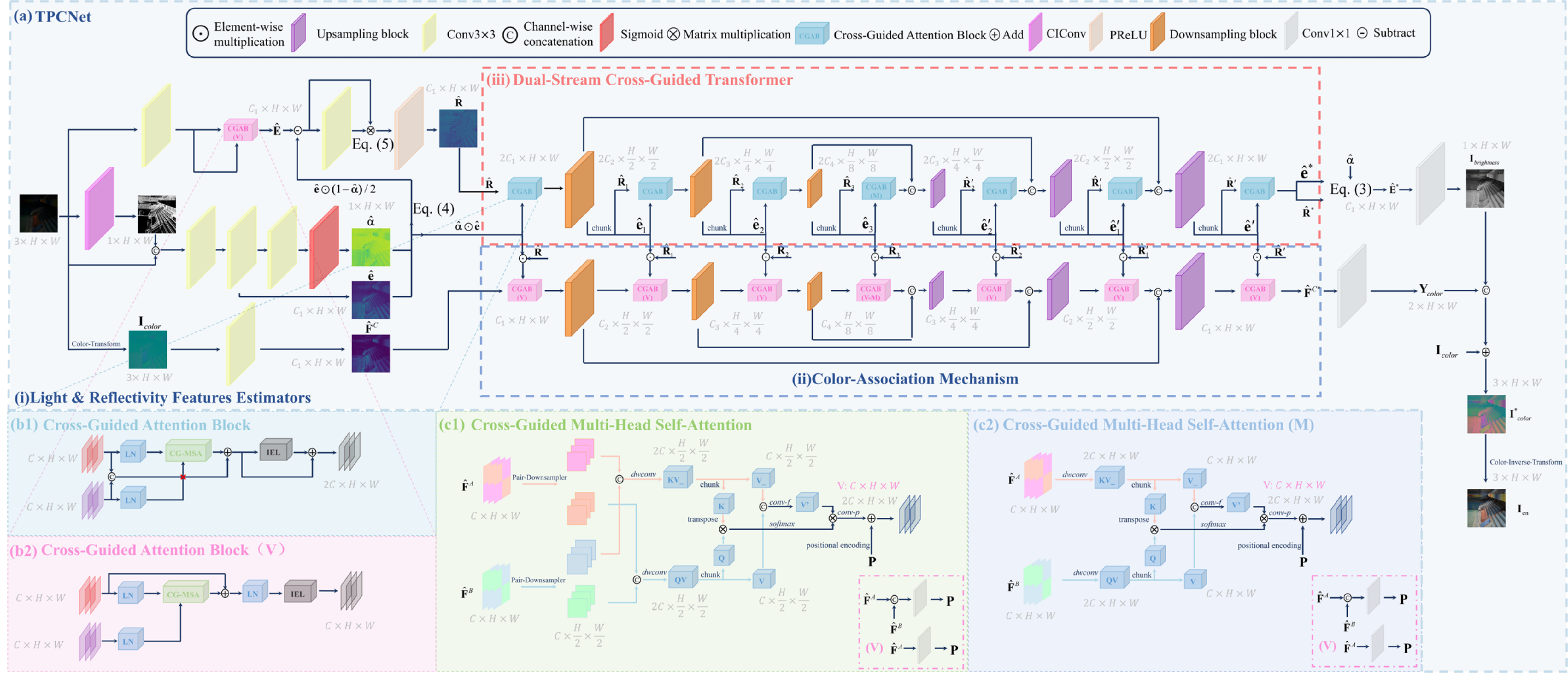}
   \caption{(a) Overview of our proposed TPCNet, which contains a Light \& Reflectivity Features Estimator (i), the Color-Association Mechanism (CAM) (ii), and the Dual-Stream Cross-Guided Transformer (iii). (b1) The Cross-Guided Attention Blocks (CGABs) are comprised of a normalization layer (LN), a Cross-Guided Multi-Head Self-Attention (CG-MSA) module, and an IEL module \cite{6}. (b2) CGAB (V) with variation in skip connection. (c1)-(c2) Detailed structure of the CG-MSA and its variation.}
   \label{fig:2}
\end{figure*}

%% file: sec/3_method.tex
% \vspace{-2mm}
\section{Method}
% \vspace{-2mm}
\label{sec:method}
Figure \ref{fig:2} displays the specific architecture of our method. As illustrated in Fig. \ref{fig:2}(a), our TPCNet is composed of the LFE, RFE, DCGT, and CAM modules, where DCGT is designed as the core component of the enhancer, comprising the basis units, namely Cross-Guided Attention Blocks (CGABs). The latter, which include variants in skip connections, positional encoding, and feature crossover, such as, CGAB (V), CGAB (V-M), and CGAB (M), are composed of a normalization layer, a CG-MSA module, and an IEL module \cite{6} similar to the Gated Forward Network. The CGAB(V) represents the variant that is different in positional encoding and skip connections, and CGAB (M) stands for the CG-MSA module with the pair-downsample removed.

\subsection{Triple Physical Constraint Framework}

To introduce our TPC concept, we start from the image formation model based on the Kubelka-Munk theory \cite{32}
\begin{equation}
  E(\lambda ,\mathbf{x}) = e(\lambda ,\mathbf{x})({(1 - {\rho _f}(\mathbf{x}))^2}{R_\infty }(\lambda ,\mathbf{x}) + {\rho _f}(\mathbf{x})),
  \label{eq:1}
\end{equation}
where $E(\lambda ,\mathbf{x})$  stands for the spectrum of light reflected from an object in the viewing direction, $\lambda$ and $\mathbf{x}$ refer to the wavelength of the light and spatial location on the image plane, respectively, $e(\lambda ,\mathbf{x})$ denotes the spectrum of the light source, ${R_\infty }(\lambda ,\mathbf{x})$ is the material reflectivity, and ${\rho _f}(\mathbf{x})$ represents the specular reflection coefficient. Following the Retinex-based image formation approximation \cite{3}, we ignore the effect of the wavelength to simplify \cref{eq:1}. But for ${\rho _f}(\mathbf{x})$, the Retinex theory \cite{2,4} assumes the objects to be Lambertian and that ${\rho _f}(\mathbf{x}) \approx 0$, whereas we view ${\rho _f}(\mathbf{x})$ as a small quantity that cannot be ignored.

Then, by expanding ${(1 - {\rho _f}(\mathbf{x}))^2}$ as a Taylor series and omitting higher-order term, we reformulate \cref{eq:1} as
\begin{equation}
  E(\mathbf{x}) = (1 - 2{\rho _f}(\mathbf{x}))e(\mathbf{x}){R_\infty }(\mathbf{x}) + {\rho _f}(\mathbf{x})e(\mathbf{x}).
  \label{eq:2}
\end{equation}
Subsequently, by setting $1 - 2{\rho _f}(\mathbf{x}) = \alpha (\mathbf{x})$, \cref{eq:2} can be written as
\begin{equation}
E(\mathbf{x}) = \underbrace {\alpha (\mathbf{x})e(\mathbf{x})}_{L(\mathbf{x})}{R_\infty }(\mathbf{x}) + \underbrace {(1 - \alpha (\mathbf{x}))e(\mathbf{x})}_{\bar L(\mathbf{x})}/2,
  \label{eq:3}
\end{equation}
where $L(\mathbf{x}) = \alpha (\mathbf{x})e(\mathbf{x})$ can be regarded as an illumination map, while $\bar L(\mathbf{x}) = (1 - \alpha (\mathbf{x}))e(\mathbf{x})$ is its complementary illumination map, and both satisfy the constraint
\begin{equation}
L(\mathbf{x}) + \bar L(\mathbf{x}) = e(\mathbf{x}),
  \label{eq:4}
\end{equation}
By transforming \cref{eq:3}, we obtain another constraint as
\begin{equation}
{R_\infty }(\mathbf{x}) = \frac{{E(\mathbf{x}) - \bar L(\mathbf{x})/2}}{{L(\mathbf{x})}},
  \label{eq:5}
\end{equation}
Equations (\ref{eq:3}) to (\ref{eq:5}) constitute a TPC for detection of the reflected light, the illumination from the light source, and the material reflectivity characteristic, respectively. This TPC thus provides a heuristic guidance for network design.

\subsection{Overview of the Triple Physical Constraint Network}
To endow the network with TPC (TPCNet), the computational graph can be formulated as:
\begin{equation}
\begin{split}
\psi (\mathbf{I}) &\xrightarrow{}(\mathbf{\hat e},\bm{\hat{\upalpha}} ) 
         \xrightarrow{\;\rm Eq.~(\ref{eq:4})\;}(\hat {\mathbf{L}},\hat {\bar {\mathbf{L}}})\xrightarrow{}\zeta(\hat {\mathbf{L}},\hat {\bar {\mathbf{L}}})\xrightarrow{} \hat {\mathbf{R}} \\
        \hat {\mathbf{R}}&\xrightarrow{}\cal {E}\rm(\hat {\mathbf{R}},\hat {\mathbf{L}})\xrightarrow{}({\hat {\mathbf{R}^*}},{\hat {\mathbf{e}}^*})\xrightarrow[\;add\ \bm{\hat{\upalpha}}\;]{\;\mathrm{Eq.~(\ref{eq:3})}\;}\hat {\mathbf{E}},
\end{split}
\label{eq:6}
\end{equation}
where $\psi$ refers to the LFE, $\zeta$ denotes the RFE and $\cal {E}$ stands for the enhancer. Taking low-light images $\mathbf{I} \in {{\cal R}^{3 \times H \times W}}$ as input, $\psi$ estimates the initial light features $\hat {\mathbf{e}} \in {{\cal R}^{C \times H \times W}}$ and corresponding weight $\bm{\hat{\upalpha}} \in {{\cal R}^{1 \times H \times W}}$ with value [0-1]. Subsequently, substituting the $\hat {\mathbf{e}}$ and $\bm{\hat{\upalpha}}$ into \cref{eq:4} as the first constraint, we can calculate the complementary illumination map $\hat {\bar {\mathbf{L}}} \in {{\cal R}^{C \times H \times W}}$. Then $\mathbf{I}$ and $\hat {\bar{\mathbf{L}}}$ are fed into $\zeta$ with the \cref{eq:5} constraint to produce the reflectivity feature $\hat R \in {{\cal R}^{C \times H \times W}}$. Ultimately, the enhanced ${\hat {\mathbf{R}}^*} \in {{\cal R}^{C \times H \times W}}$ and ${\hat {\mathbf{e}}^*} \in {{\cal R}^{C \times H \times W}}$ are calculated by a four-scale U-shaped architecture $\cal {E}$ and we substitute ${\hat {\mathbf{R}}^*}$, ${\hat {\mathbf{e}}^*}$ and $\bm{\hat{\upalpha}}$ into \cref{eq:3} to obtain the reflected-light feature ${\hat {\mathbf{E}}^*} \in {{\cal R}^{C \times H \times W}}$, thus realizing the third physical constraint on the network.

However, since the effect of wavelength has been ignored in our model, directly mapping ${\hat {\mathbf{E}}^*}$ to ${\mathbf{I}_{{en}}} \in {{\cal R}^{3 \times H \times W}}$ may result in color deviation or overexposure. To overcome this limitation, we introduce a multi-scale color-association mechanism. First, we convert the RGB color space of $\mathbf{I}$ into another color space ${\mathbf{I}_{color}} \in {{\cal R}^{3 \times H \times W}}$ (\eg, YCbCr \cite{34}, HSV \cite{5}, HVI \cite{6}, \textit{etc.}) to realize the separation of brightness and color information. Subsequently, the color features ${{\hat {\mathbf{F}}}}^C \in {{\cal R}^{C \times H \times W}}$ are extracted from ${\mathbf{I}_{color}}$, and are then fed into the DCGTs to obtain the enhanced color feature ${\hat{\mathbf{F}}}^{C^*} \in {{\cal R}^{C \times H \times W}}$ with the multi-scale ${\hat {\mathbf{R}}^*} \odot {\hat {\mathbf{e}}^*}$ information as feature guidance. Next, the ${{\mathbf{F}}^C}^*$ and ${\hat {\mathbf{E}}^*}$ are mapped to the color information ${\mathbf{Y}_{color}} \in {{\cal R}^{2 \times H \times W}}$ and brightness ${\mathbf{I}_{brightness}} \in {{\cal R}^{1 \times H \times W}}$, respectively, and we concatenate ${{\mathbf{Y}}_{color}}$ and ${\mathbf{I}_{brightness}}$  for each pixel along the channel dimension to acquire the ${{\mathbf{I}}^*}_{color} \in {{\cal R}^{3 \times H \times W}}$ and introduce the residual connection ${\mathbf{I}^*}_{color} = {\mathbf{I}^*}_{color} + {\mathbf{I}_{color}}$. Finally, the ${\mathbf{I}^*}_{color}$ is transformed back into the ${\mathbf{I}_{{\rm{en}}}}$.

The architecture of $\psi$ is shown in Fig. (\ref{fig:2}), where $\psi$ first extracts the invariance properties $W$ of the illumination intensity from $\mathbf{I}$ by a CIConv \cite{1_7}, and a $conv3\!\times\!3$ (convolution with kernel size = 3) is used to fuse the concatenation of $\mathbf{I}$ and $W$. Since the fused feature possesses rich semantic contextual information, we utilize a $conv3\!\times\!3$ operation to generate $\hat {\mathbf{e}}$, and the corresponding weight $\bm{\hat{\upalpha}}$ is calculated by another $conv3\!\times\!3$ followed by Sigmoid activation.

The composition of $\zeta$ is exhibited in Fig. \ref{fig:2}(a). First, $\zeta$ employs a $conv3\!\times\!3$ followed by the CGAB to estimate $\hat {\mathbf{E}}$ from ${\mathbf{I}}$ and calculates $\hat {\mathbf{E}} - \frac{{\hat {\bar {\mathbf{L}}}}}{2}$. Subsequently, $\hat {\mathbf{E}} - \frac{{\hat {\bar {\mathbf{L}}}}}{2}$ is fed into another $conv3\!\times\!3$ to produce $\hat {\mathbf{L}}^{'}$ where $\hat {\mathbf{L}}^{'} \odot \hat {\mathbf{L}} = 1$. To enhance the physical constraints, we substitute $\hat {\mathbf{E}} - \frac{{\hat {\bar {\mathbf{L}}}}}{2}$ and $\hat {\mathbf{L}}^{'}$ into \cref{eq:5} to compute $\hat {\mathbf{R}}$.

\noindent\textbf{Discussion.} \textbf{(i)} Compared with the previous deep learning methods based on the Retinex theory \cite{3,8,9,10,11_39} that view the $\hat {\mathbf{R}}$ and $\hat {\mathbf{L}}$ as independent estimations and employ deep convolutional neural networks to estimate separately or to estimate $\hat {\mathbf{L}}$ only, the TPC framework regards the $\hat {\mathbf{R}}$ and $\hat {\mathbf{L}}$ as the estimations with mathematical connection, and then the $\hat {\mathbf{R}}$ can be calculated by $(\hat {\mathbf{E}} - \frac{{\hat {\bar {\mathbf{L}}}}}{2}) \odot \hat {\mathbf{L}}^{'}$. This framework utilizes the inherent relationships between $\hat {\mathbf{R}}$ and $\hat {\mathbf{L}}$ for self-constraint without requiring complicated loss functions, which can be effective in decreasing training difficulty. Moreover, the calculation process $\hat R = (\hat {\mathbf{E}} - \frac{{\hat {\bar {\mathbf{L}}}}}{2}) \odot \hat {\mathbf{L}}^{'}$ possesses some advantages. From one perspective, $(\hat {\mathbf{E}} - \frac{{\hat {\bar {\mathbf{L}}}}}{2})$ can decrease the estimation bias to a certain extent by computing the difference between $\hat {\mathbf{E}}$ and $\frac{{\hat {\bar {\mathbf{L}}}}}{2})$. From another perspective, $\hat R = (\hat {\mathbf{E}} - \frac{{\hat {\bar {\mathbf{L}}}}}{2}) \odot \hat {\mathbf{L}}^{'}$ avoids directly dividing $(\hat {\mathbf{E}} - \frac{{\hat {\bar {\mathbf{L}}}}}{2})$ by $\hat {\mathbf{L}}$, which can enlarge the deviation \cite{12} and can produce a more robust $\hat {\mathbf{R}}$.

\noindent\textbf{(ii)} Inspired by the Retinex theory \cite{4}, we employ $\hat {\mathbf{R}} \odot \hat {\mathbf{e}}$, a rough light-up feature, as an initial guidance to produce the corresponding color features in our CAM. However, multiplying two estimated values directly might intensify the deviation due to the existence of estimation bias, and the expanded deviation will further contaminate the subsequent output by forward propagation, affecting model performance. To address these limitations, we adopt multi-scale light-up features and recalculate each scale-guided feature by employing enhanced light-up instead of the skip connections from the initial $\hat {\mathbf{R}} \odot \hat {\mathbf{e}}$ to the next scale, thus diminishing the deviation accumulation.
% \vspace{-1mm}
\subsection{Dual-Stream Cross-Guided Transformer}
% \vspace{-1mm}
Recently, due to their superiority in capturing long-range dependencies, transformers have been increasingly exploited in image enhancement or restoration. Their computing mechanism can be flexibly used to calculate the correlation between features and to realize cross guidance. Some transformer-based guided mechanism, like Retinexformer \cite{12}, realize feature guidance by value elements $\mathbf{V}$ multiplied with ${\mathbf{F}_{lu}}$; others, like CIDNet \cite{6}, calculate $\mathbf{KV}$ elements and $\mathbf{Q}$ from two different features using separate learnable layers, and then realize cross guidance based on a self-attention mechanism (SAM). These guidance mechanisms have inherent limitations. Facing different features, the separate learnable layers with distinct learning abilities can cause attention bias, and the output can be affected by the portion that possesses stronger learning capabilities. To address this limitation, we design a DCGT as the enhancer ${\cal E}$.

\noindent\textbf{Network Structure.} As illustrated in Fig. \ref{fig:2}(a), DCGT exploits a four-scale U-Structure. Initial $\hat {\mathbf{L}}$, $\hat {\mathbf{R}}$ and ${\hat {\mathbf{F}}^C}$, are input into DCGT. During downsampling, $\hat {\mathbf{L}}$ and $\hat {\mathbf{R}}$ are processed by a CGAB and a downsampling block which consists of $conv3\!\times\!3$ (for channel conversion), bilinear interpolation, and $\mathrm{PReLU}$ activation, to generate $\hat {\mathbf{F}}_i^{RL} \in {{\cal R}^{2{C_i} \times \frac{H}{{{2^i}}} \times \frac{W}{{{2^i}}}}}$ where $i$ = 1, 2, 3. Then $\hat {\mathbf{F}}_i^{RL}$ is partitioned into ${\hat {\mathbf{R}}_i}$ and ${\hat {\mathbf{e}}_i}$ along the channel dimension; ${\hat {\mathbf{R}}_i}$ and ${\hat {\mathbf{e}}_i}$ as the new inputs are fed into the CGAB. For the downsampling color-association portion, ${\hat {\mathbf{F}}^C}$ and $\hat {\mathbf{R}} \odot \hat {\mathbf{L}}$ undergo a CGAB and the downsampling block to produce the $\hat {\mathbf{F}}_i^C \in {{\cal R}^{{C_i} \times \frac{H}{{{2^i}}} \times \frac{W}{{{2^i}}}}}$. As the guidance for $\hat {\mathbf{F}}_i^C$, ${\hat {\mathbf{R}}_i} \odot {\hat {\mathbf{e}}_i}$ is applied in the next CGAB. Then ${\hat {\mathbf{R}}_3}$ and ${\hat {\mathbf{e}}_3}$ pass through CGAB (M) while $\hat {\mathbf{F}}_3^C$ and ${\hat {\mathbf{R}}_3} \odot {\hat {\mathbf{e}}_3}$ are fed into CGAB (V-M), respectively. Subsequently, a symmetrical structure is designed for upsampling. The upsampling block is utilized to fuse skip connections and to upscale the features, which is composed of $conv3\!\times\!3$ (for channel conversion), $conv1\!\times\!1$ (for connection fusion) bilinear interpolation, and $\mathrm{PReLU}$ activation. The color space of TPCNet is Ycbcr, and the results of ablation experiments shown in Table \ref{tab:3} verify the effect of different color spaces.

\begin{table*}[!htb]
\caption{Quantitative comparison with LOL v2 \cite{29} and VILNC-indoor \cite{25} datasets using different evaluation methods (PSNR$\uparrow$, SSIM$\uparrow$, and LPIPS$\downarrow$). The best and second-best performances are highlighted in \textcolor{red}{red} and \textcolor{cyan}{cyan}, respectively. The FLOPs is calculated on a single 256$\times$256 image.}

\label{tab:1}
\vspace{-1mm}
\centering
\resizebox{\textwidth}{!}{%
\begin{tabular}{cc|cc|ccc|ccc|ccc}
\noalign{\hrule height 1.6pt} % 横线粗细1pt
\multicolumn{2}{c|}{\multirow{2}{*}{\textbf{Methods}}} & \multicolumn{2}{c|}{\textbf{Complexity}} & \multicolumn{3}{c|}{\textbf{LOL-v2-Real}} & \multicolumn{3}{c|}{\textbf{LOL-v2-Synthetic}} & \multicolumn{3}{c}{\textbf{VILNC-Indoor}} \\
&& Params/M & FLOPs/G & PSNR$\uparrow$ & SSIM$\uparrow$ & LPIPS$\downarrow$ & PSNR$\uparrow$ & SSIM$\uparrow$ & LPIPS$\downarrow$ & PSNR$\uparrow$ & SSIM$\uparrow$ & LPIPS$\downarrow$ \\
% \hline
\specialrule{1pt}{1pt}{1pt} % 第一个参数：粗细, 后两个：上下间距
\multicolumn{2}{c|}{KinD \cite{56}}& 8.02 & 34.99 & 17.544 & 0.669 & 0.375 & 18.320 & 0.796 & 0.252 & 9.453 & 0.583 & 0.411 \\
\multicolumn{2}{c|}{RUAS \cite{57}}& 0.003 & 0.83 & 15.506 & 0.491 & 0.314 & 13.880 & 0.676 & 0.291 & 8.836 & 0.570 & 0.735 \\
\multicolumn{2}{c|}{RetinexNet \cite{3}}& 0.84 & 584.47 & 16.097 & 0.401 & 0.543 & 17.137 & 0.762 & 0.255 & 16.573 & 0.456 & 0.547 \\
\multicolumn{2}{c|}{ZeroDCE++ \cite{58}}& 0.01 & 0.61 & 12.338 & 0.443 & 0.488 & 15.871 & 0.804 & 0.217 & 15.992 & 0.434 & 0.579 \\
\multicolumn{2}{c|}{SNR-Net \cite{16}}& 4.01 & 26.35 & 21.480 & 0.849 & 0.163 & 24.140 & 0.928 & 0.056 & 23.367 & 0.765 & 0.202 \\
\multicolumn{2}{c|}{LLFormer \cite{59}}& 24.55 & 22.52 & 20.056 & 0.792 & 0.211 & 24.038 & 0.909 &0.066 & 23.593 & 0.769 & 0.313 \\
\multicolumn{2}{c|}{Retinexformer \cite{12}}& 1.61 & 15.57 & 22.794 & 0.840 & 0.171 & 25.670 & 0.930 & 0.059 & \textcolor{cyan}{23.735} & 0.782 & 0.295 \\
\multicolumn{2}{c|}{PairLIE \cite{60}}& 0.33 & 20.81 & 19.885 & 0.778& 0.317 & 19.074 & 0.794 & 0.230 & 9.255 & 0.570 & 0.681 \\
\multicolumn{2}{c|}{EnlightenGAN \cite{61}}&114.35 & 61.01 & 18.230 & 0.617 & 0.309 & 16.570 &0.734 & 0.220 & 18.976 & 0.615 & 0.354 \\
\multicolumn{2}{c|}{HVI-CIDNet \cite{6}}&1.88&7.57&\textcolor{cyan}{24.111}&\textcolor{cyan}{0.868}&\textcolor{cyan}{0.116}&25.705& \textcolor{cyan}{0.942}&0.045 & 23.656 & 0.785 & 0.303 \\
\multicolumn{2}{c|}{QuadPrior \cite{2}}& 1252.75 & 1103.2 & 20.592 & 0.811 & 0.202 & 16.108 & 0.758 & 0.114 & 11.293 & 0.552 & 0.487 \\
\multicolumn{2}{c|}{LANet \cite{62}}& 0.58 & 43.23 & 18.074 & 0.735 & 0.364 & 18.088 & 0.807 & 0.183 & 18.410 & 0.734 & 0.484 \\
\multicolumn{2}{c|}{ZERO-IG \cite{25}}& 0.087 &11.37 & 16.943 & 0.738 & 0.395& 17.610 & 0.743 & 0.410 & 12.002 & 0.595 & 0.504 \\
\multicolumn{2}{c|}{SMG \cite{36}}& 17.8 & 42.93 & 24.032 & 0.818 & 0.169 & 24.979 & 0.891& 0.092 & 23.440 & 0.688 & \textcolor{cyan}{0.212} \\
\multicolumn{2}{c|}{RQ-LLIE \cite{37}}& 13.48& 216.78& 22.371 & 0.854 & 0.142& \textcolor{cyan}{25.937} & 0.941 & \textcolor{cyan}{0.044} & 23.540 & \textcolor{red}{0.791} & 0.288 \\
\specialrule{1pt}{1.2pt}{1.2pt} % 第一个参数：粗细, 后两个：上下间距
\multicolumn{2}{c|}{\textbf{TPCNet(Ours)}}& 2.62 & 8.76 & \textcolor{red}{24.978} & \textcolor{red}{0.882} & \textcolor{red}{0.102}& \textcolor{red}{26.032}& \textcolor{red}{0.943}& \textcolor{red}{0.041}& \textcolor{red}{24.634} & \textcolor{cyan}{0.785} & \textcolor{red}{0.194} \\
\noalign{\hrule height 1.6pt} % 横线粗细1pt
\end{tabular}}
\end{table*}
\begin{figure*}[t]
\vspace{-3mm}

 \begin{minipage}[t]{0.12\linewidth}
 	\vspace{0pt}
 	\centerline{\includegraphics[width=\textwidth]{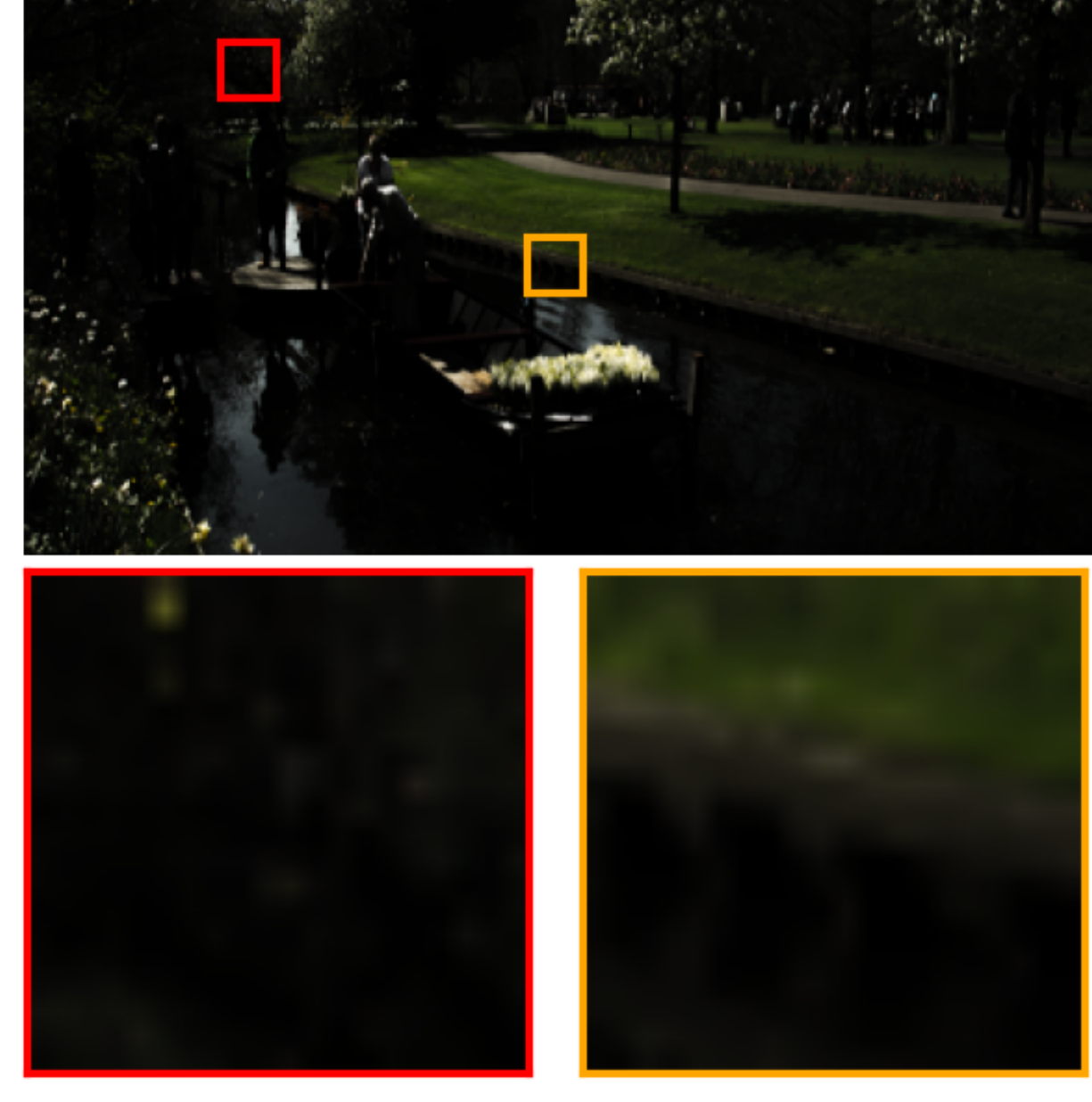}}
 	\vspace{0.2pt}
 	\centerline{\includegraphics[width=\textwidth]{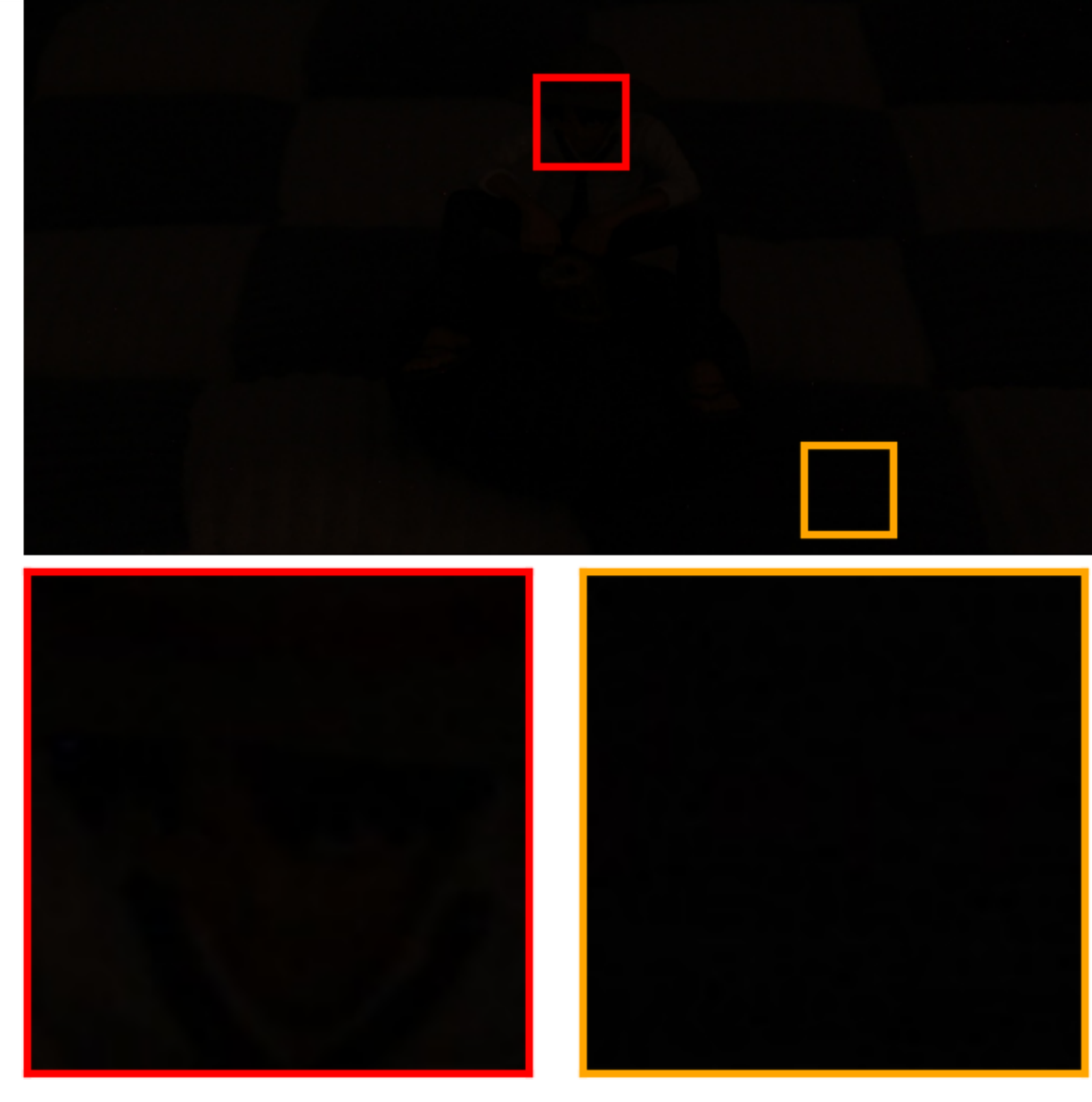}}
    \vspace{-1mm}
    \centerline{\footnotesize Input}
    \end{minipage}
 \begin{minipage}[t]{0.12\linewidth}
 	\vspace{0pt}
 	\centerline{\includegraphics[width=\textwidth]{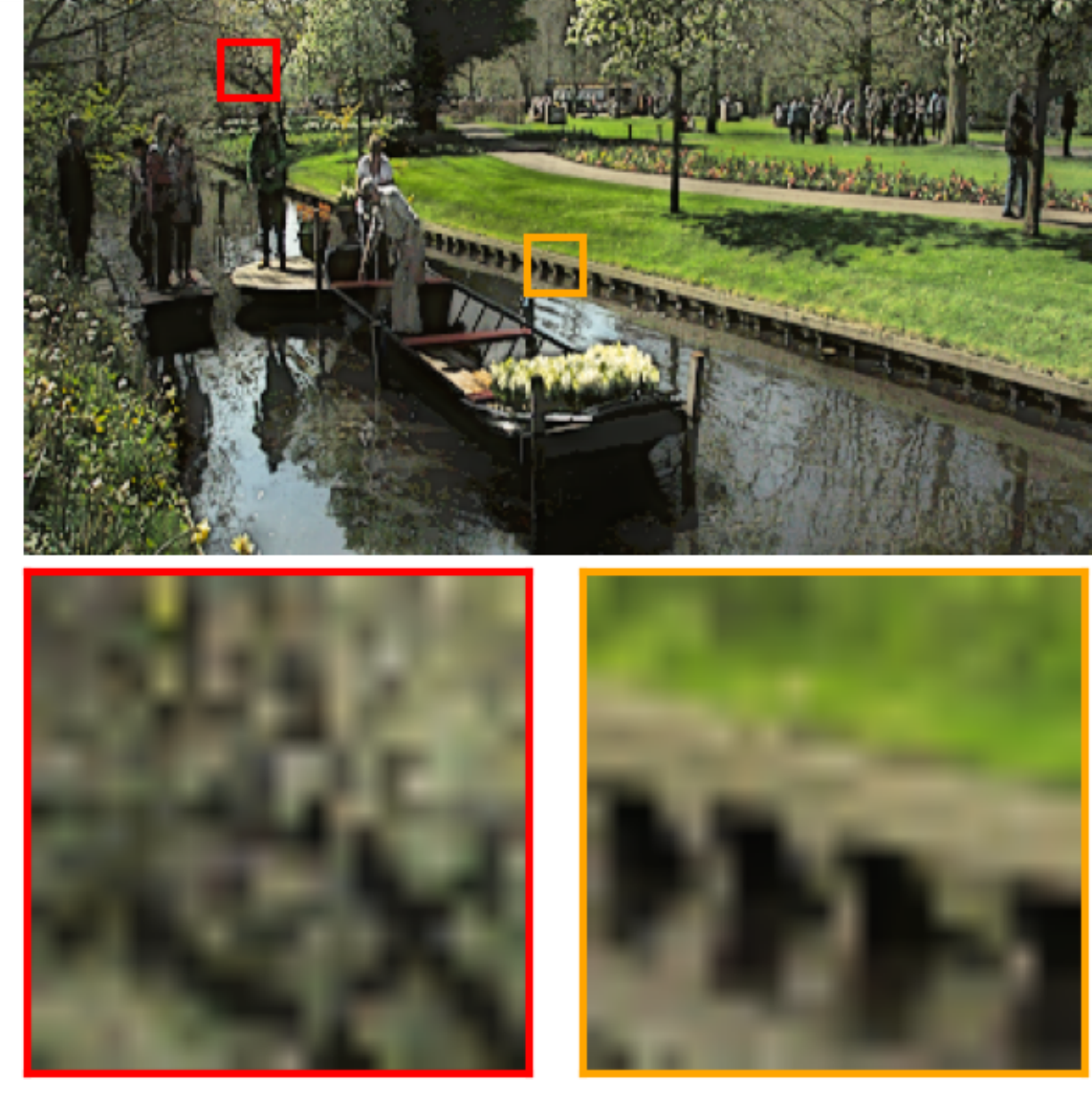}}
 	\vspace{0.2pt}
 	\centerline{\includegraphics[width=\textwidth]{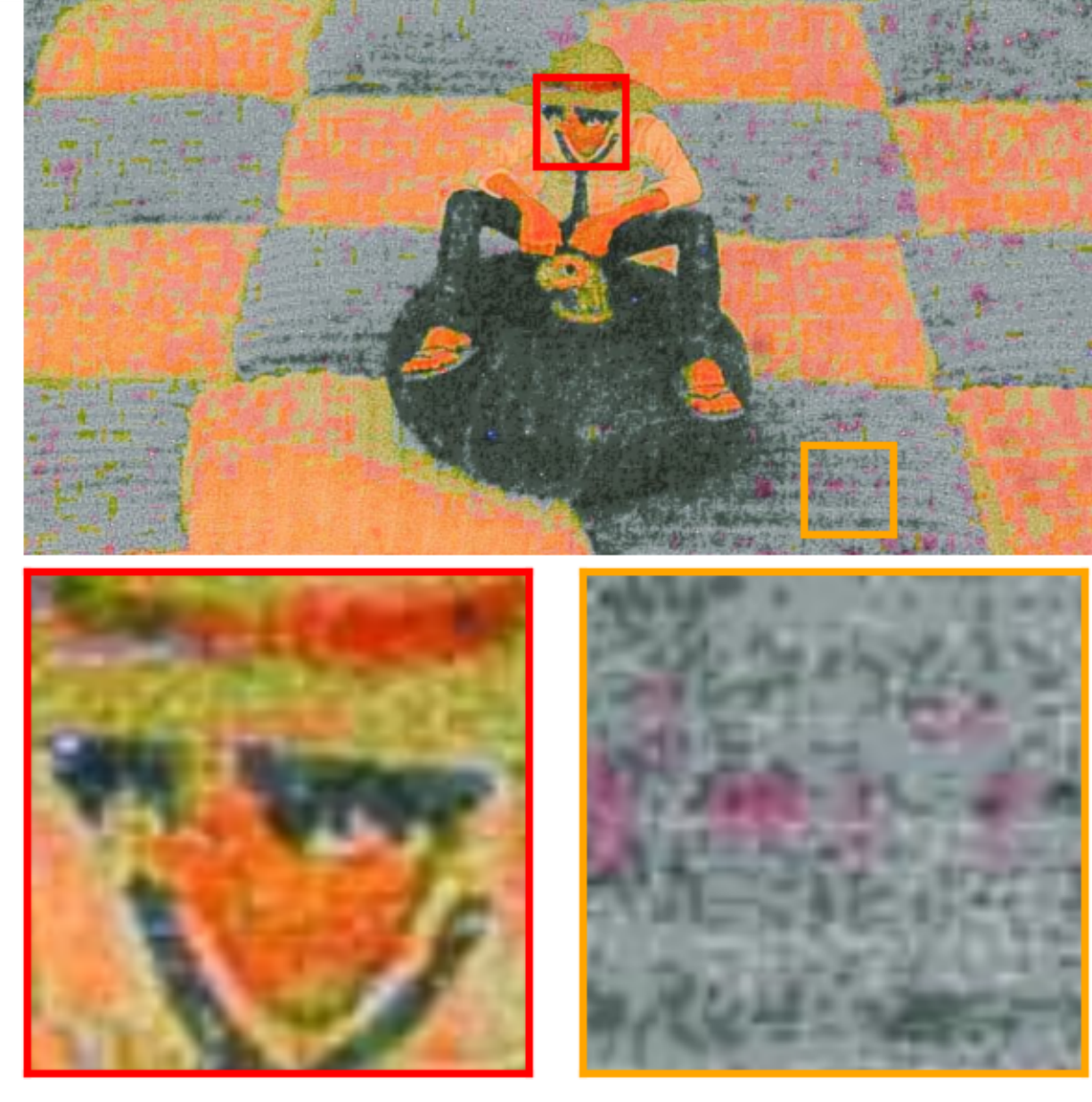}}
    \vspace{-1mm}
    \centerline{\footnotesize RetinexNet \cite{3}}
    \end{minipage}
 \begin{minipage}[t]{0.12\linewidth}
 	\vspace{0pt}
 	\centerline{\includegraphics[width=\textwidth]{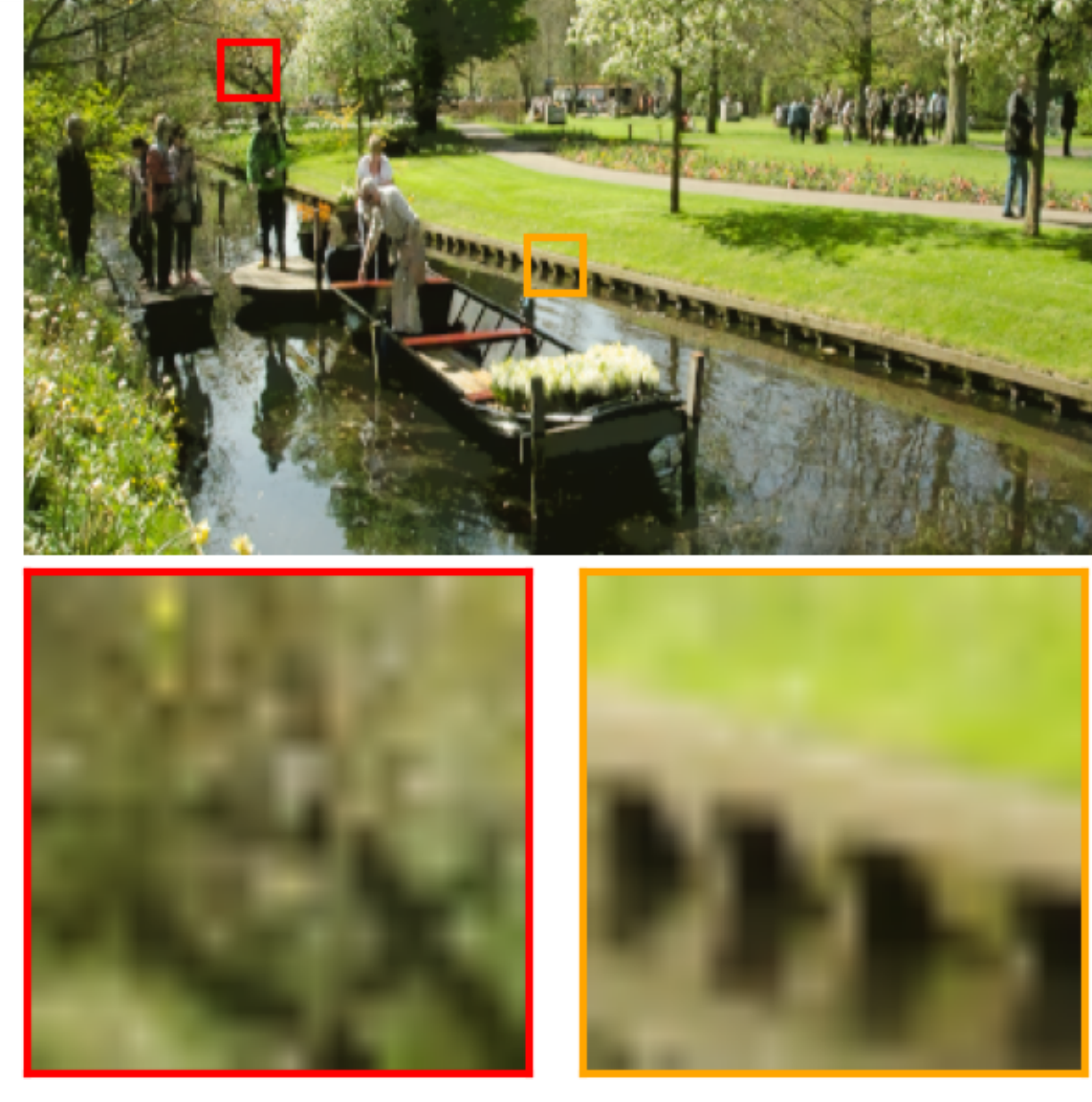}}
 	\vspace{0.2pt}
 	\centerline{\includegraphics[width=\textwidth]{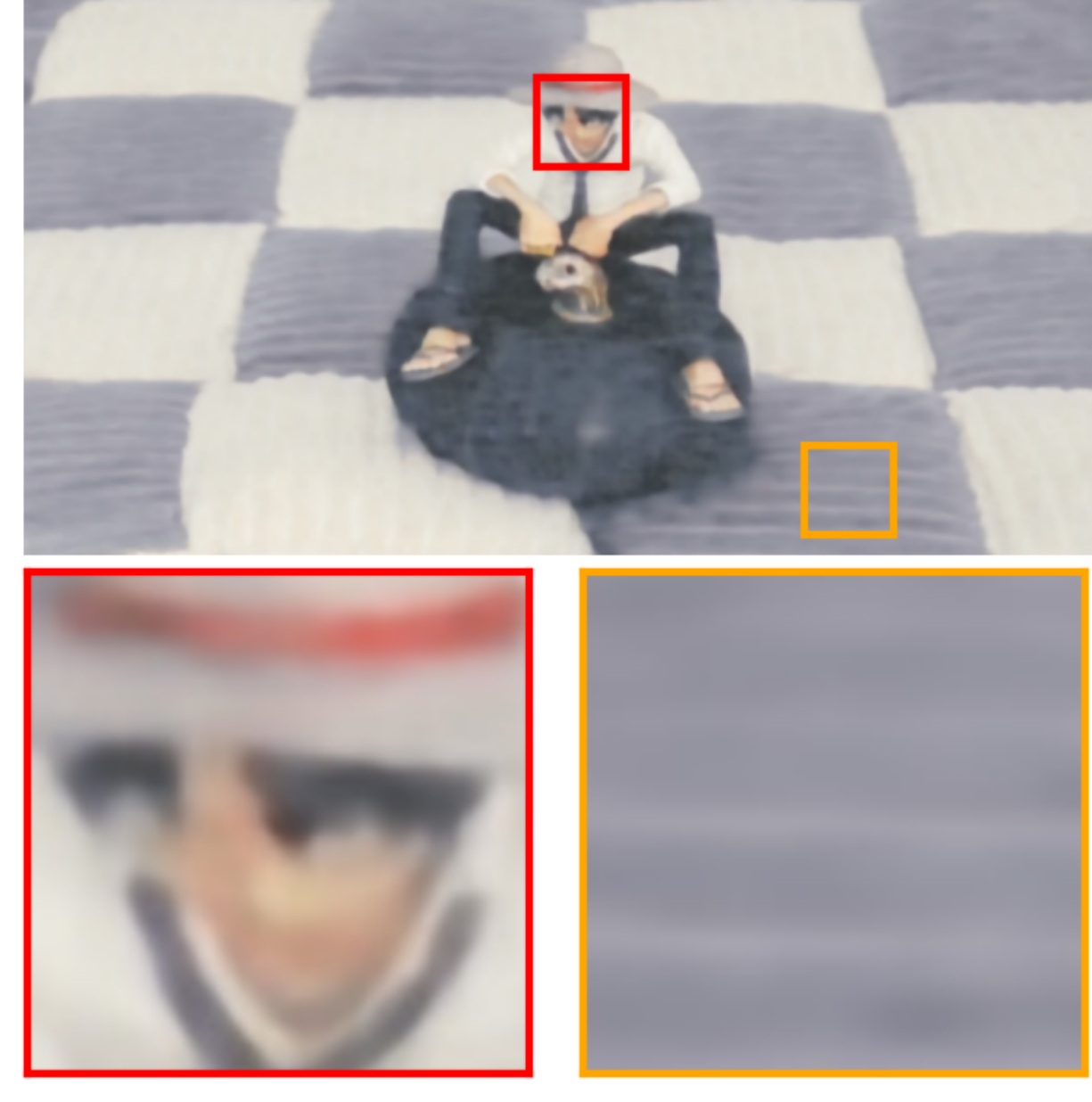}}
    \vspace{-1mm}
    \centerline{\footnotesize RQ-LLIE \cite{37}}
    \end{minipage}
 \begin{minipage}[t]{0.12\linewidth}
 	\vspace{0pt}
 	\centerline{\includegraphics[width=\textwidth]{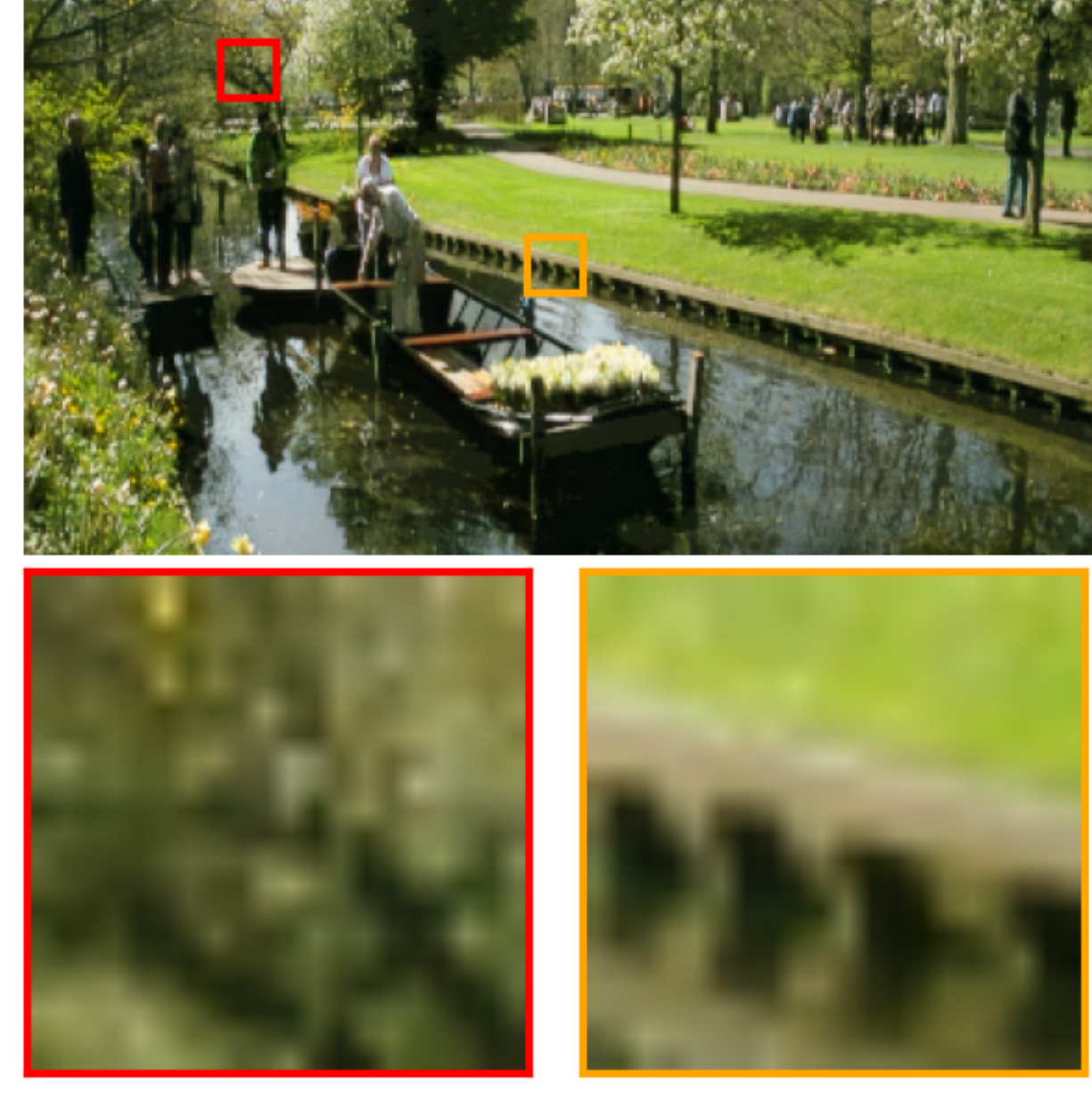}}
 	\vspace{0.2pt}
 	\centerline{\includegraphics[width=\textwidth]{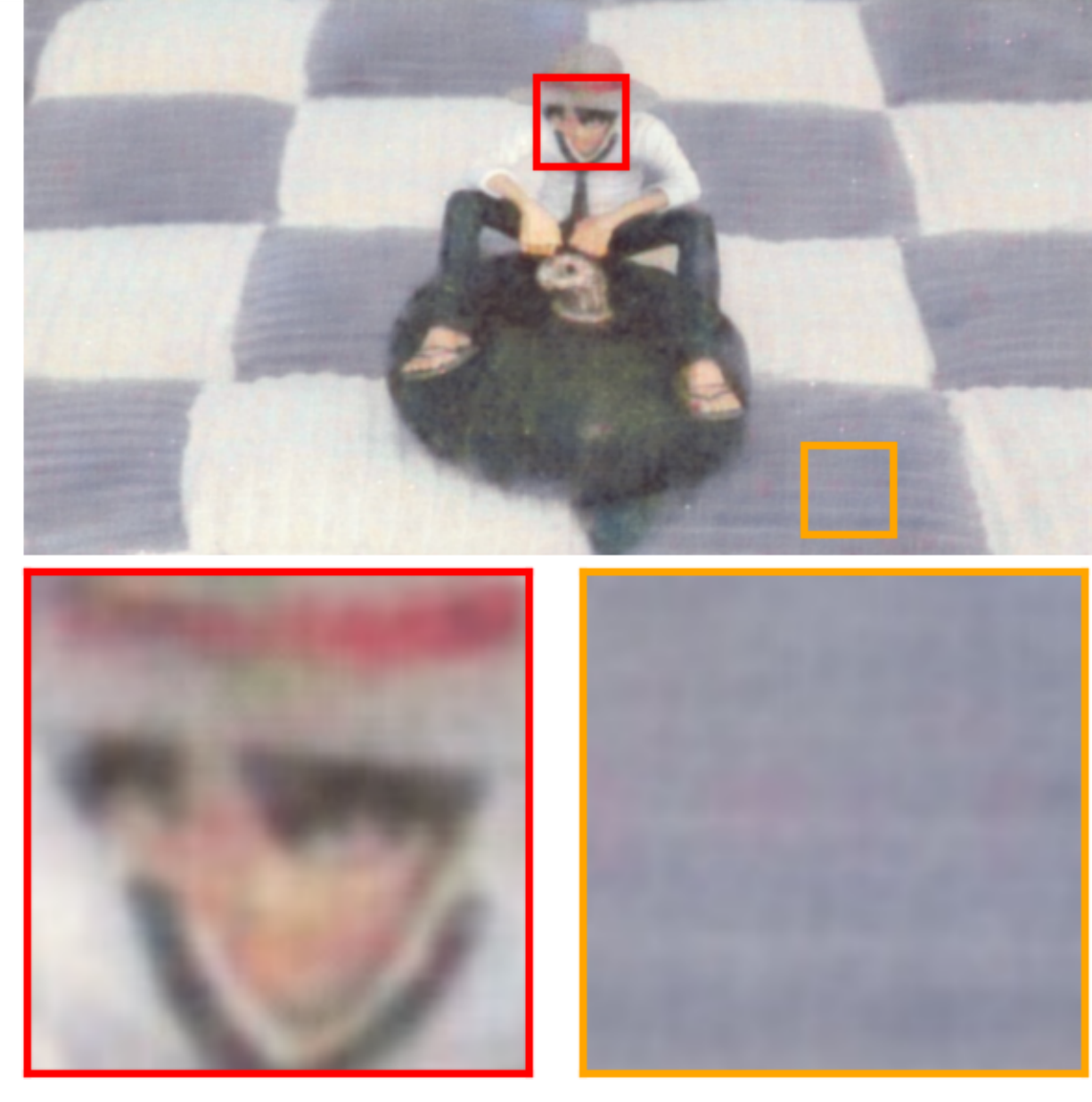}}
    \vspace{-1mm}
    \centerline{\footnotesize LLFormer \cite{59}}
    \end{minipage}
 \begin{minipage}[t]{0.12\linewidth}
 	\vspace{0pt}
 	\centerline{\includegraphics[width=\textwidth]{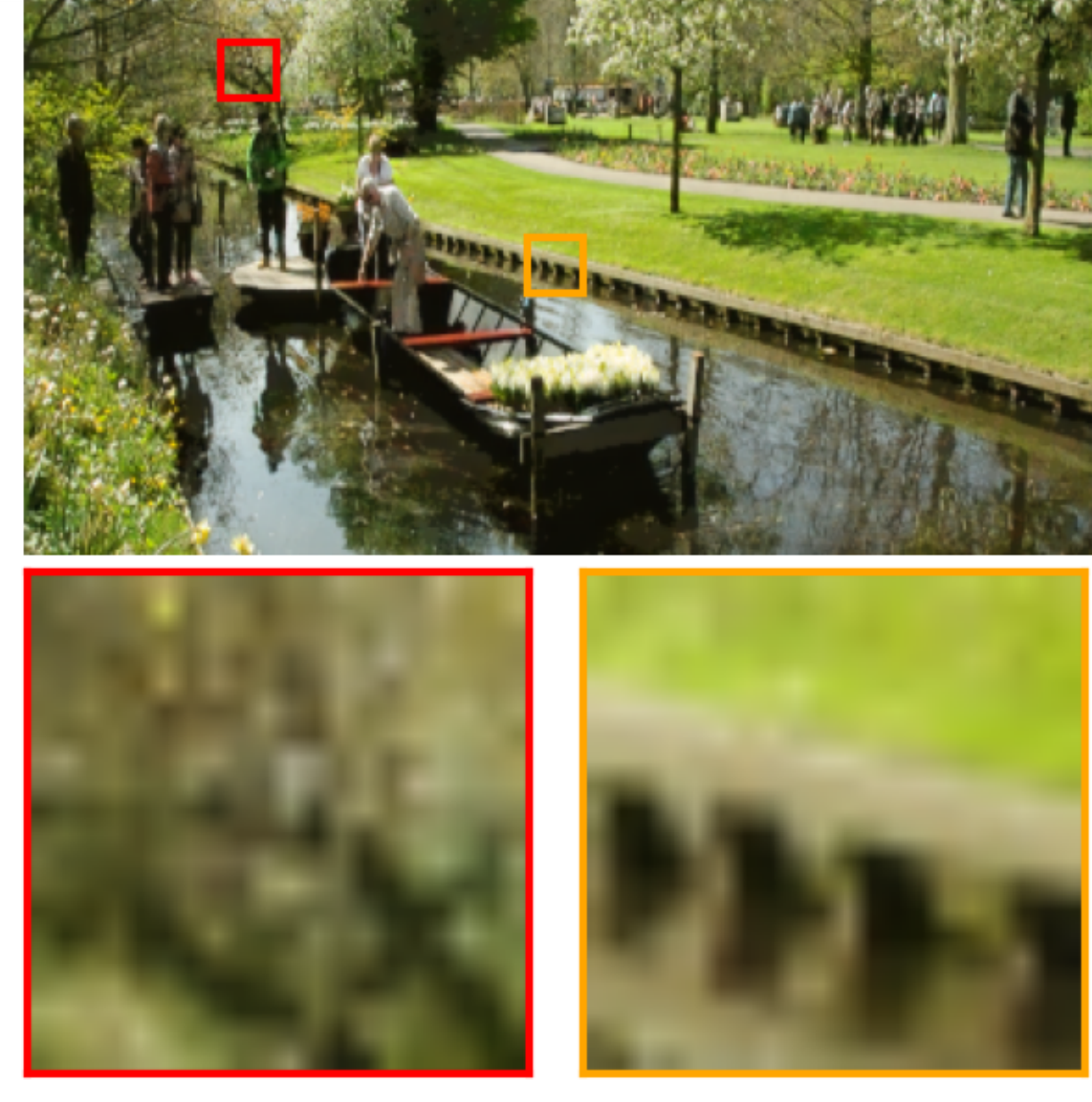}}
 	\vspace{0.2pt}
 	\centerline{\includegraphics[width=\textwidth]{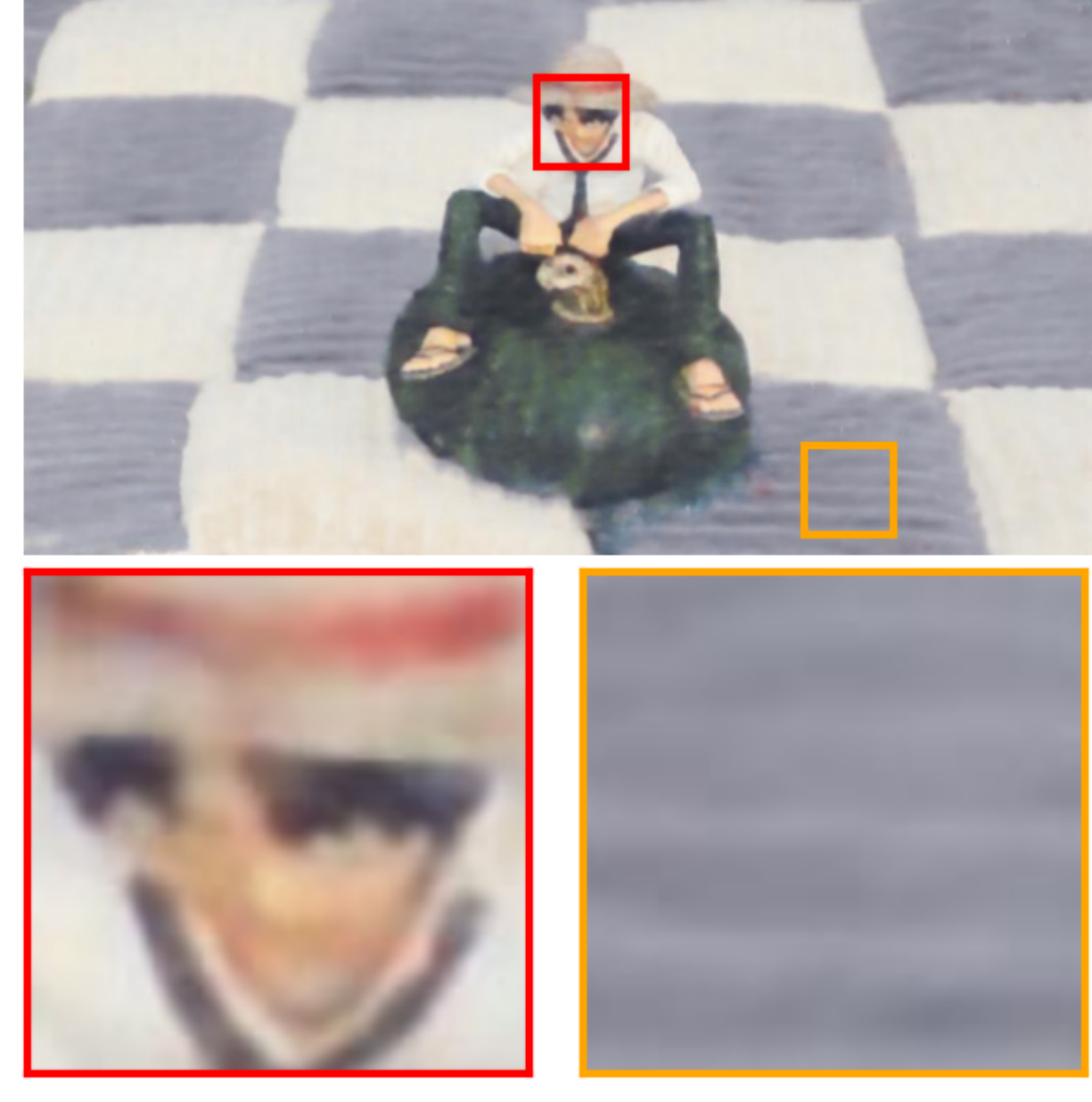}}
    \vspace{-1mm}
    \centerline{\footnotesize RetFormer \cite{12}}
    \end{minipage}
 \begin{minipage}[t]{0.12\linewidth}
 	\vspace{0pt}
 	\centerline{\includegraphics[width=\textwidth]{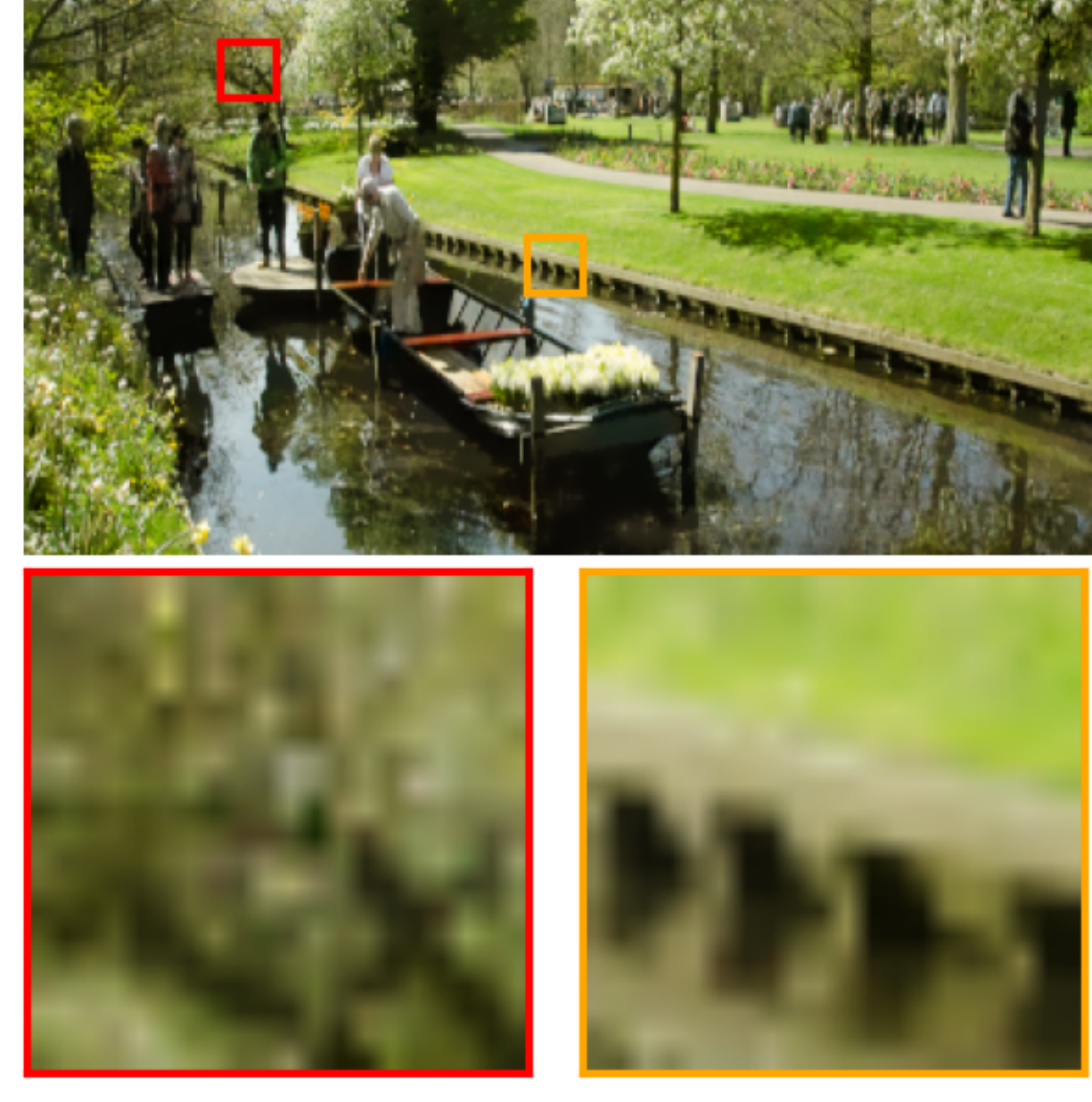}}
 	\vspace{0.2pt}
 	\centerline{\includegraphics[width=\textwidth]{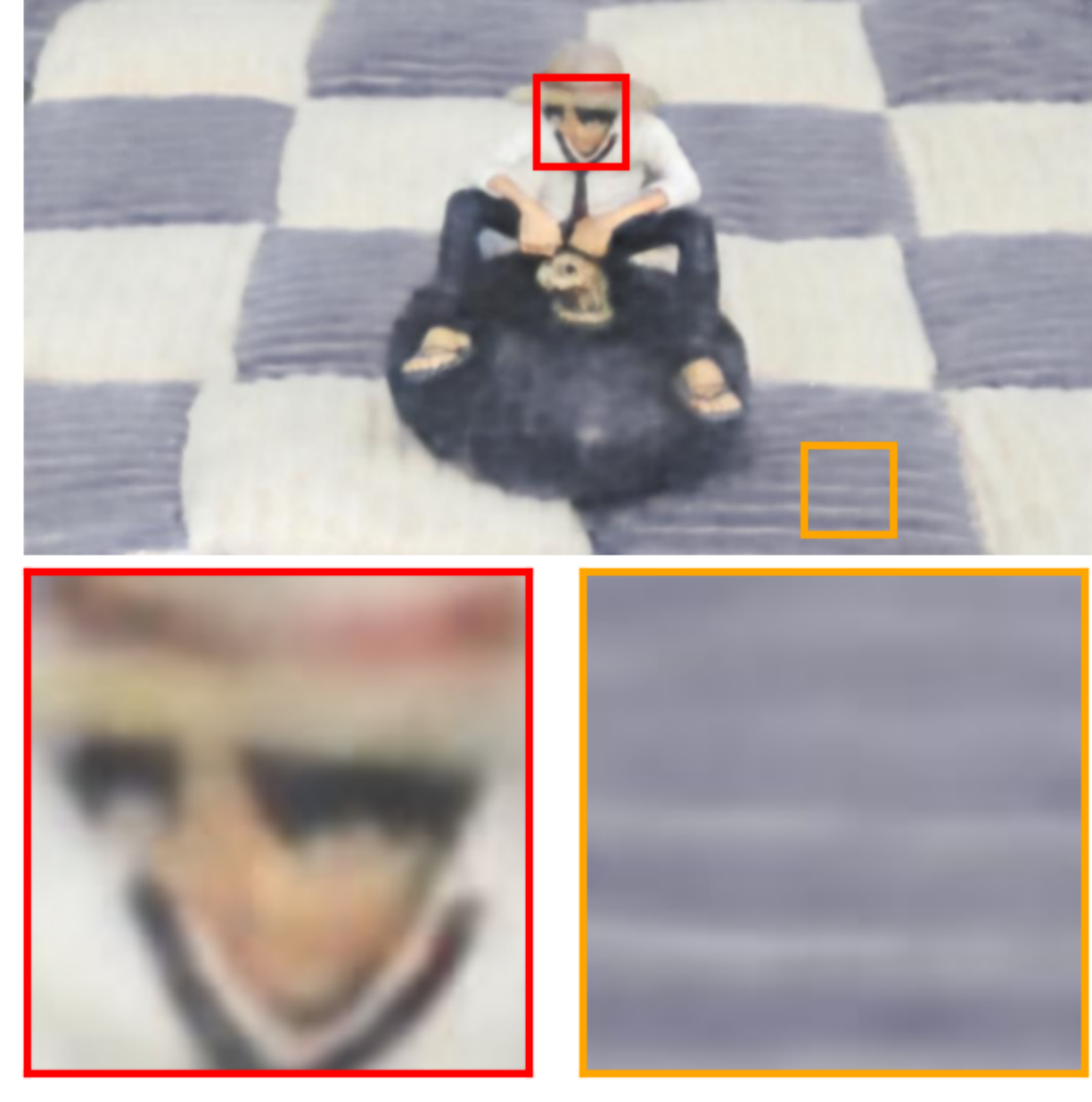}}
    \vspace{-1mm}
    \centerline{\footnotesize HVI-CIDNet \cite{6}}
    \end{minipage}
 \begin{minipage}[t]{0.12\linewidth}
 	\vspace{0pt}
 	\centerline{\includegraphics[width=\textwidth]{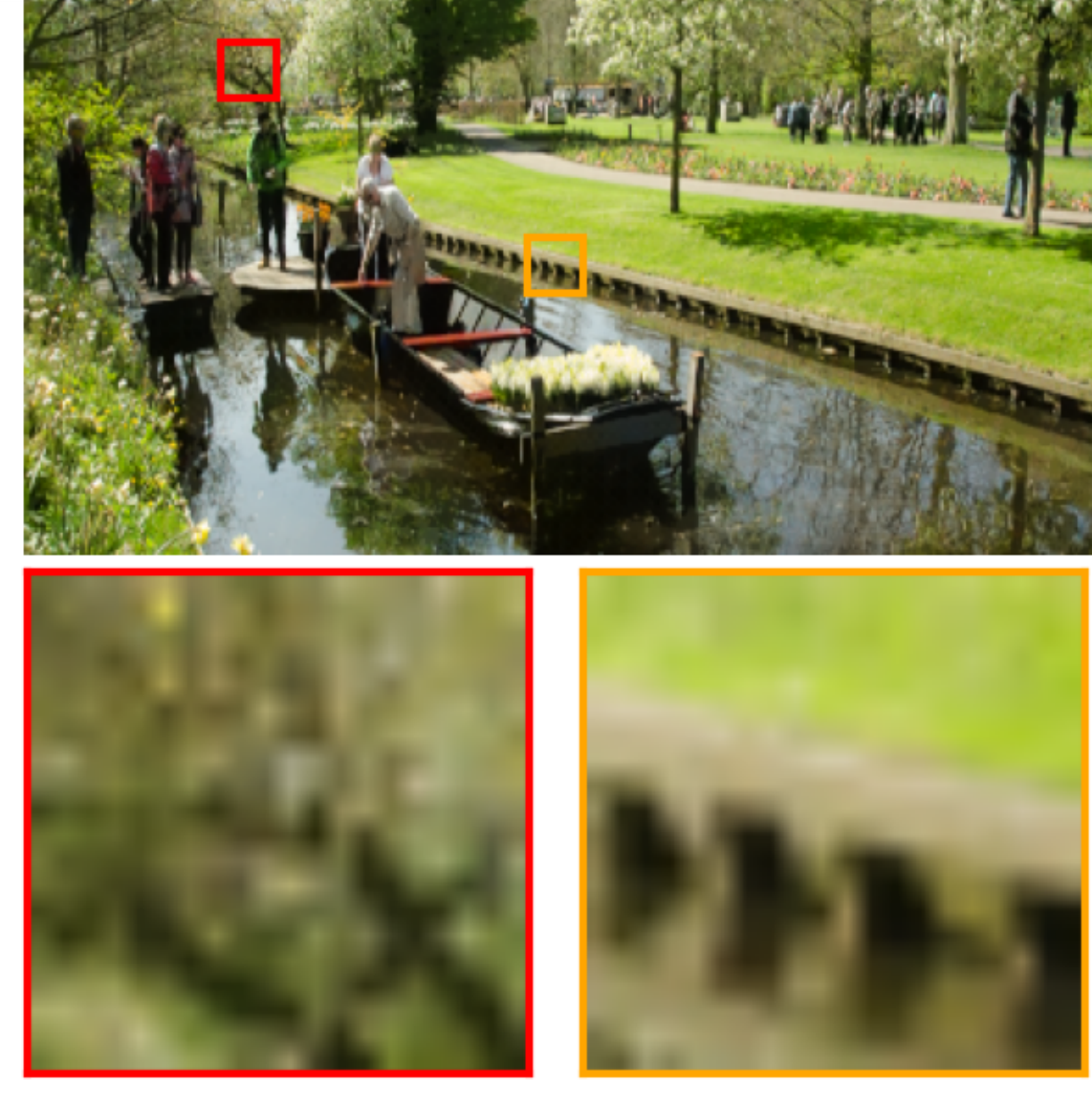}}
 	\vspace{0.2pt}
 	\centerline{\includegraphics[width=\textwidth]{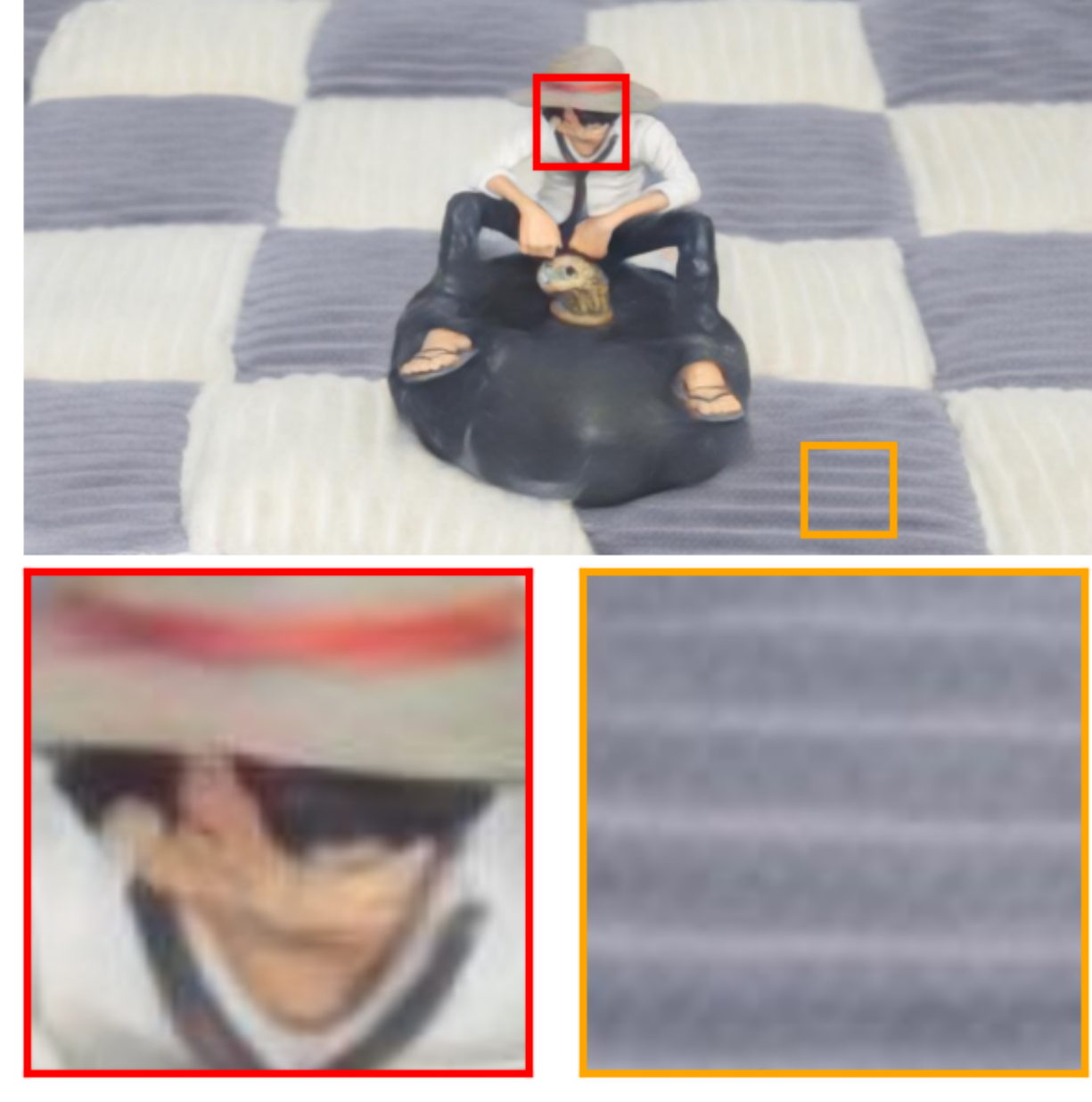}}
    \vspace{-1mm}
    \centerline{\footnotesize TPCNet (ours)}
    \end{minipage}
 \begin{minipage}[t]{0.12\linewidth}
 	\vspace{0pt}
 	\centerline{\includegraphics[width=\textwidth]{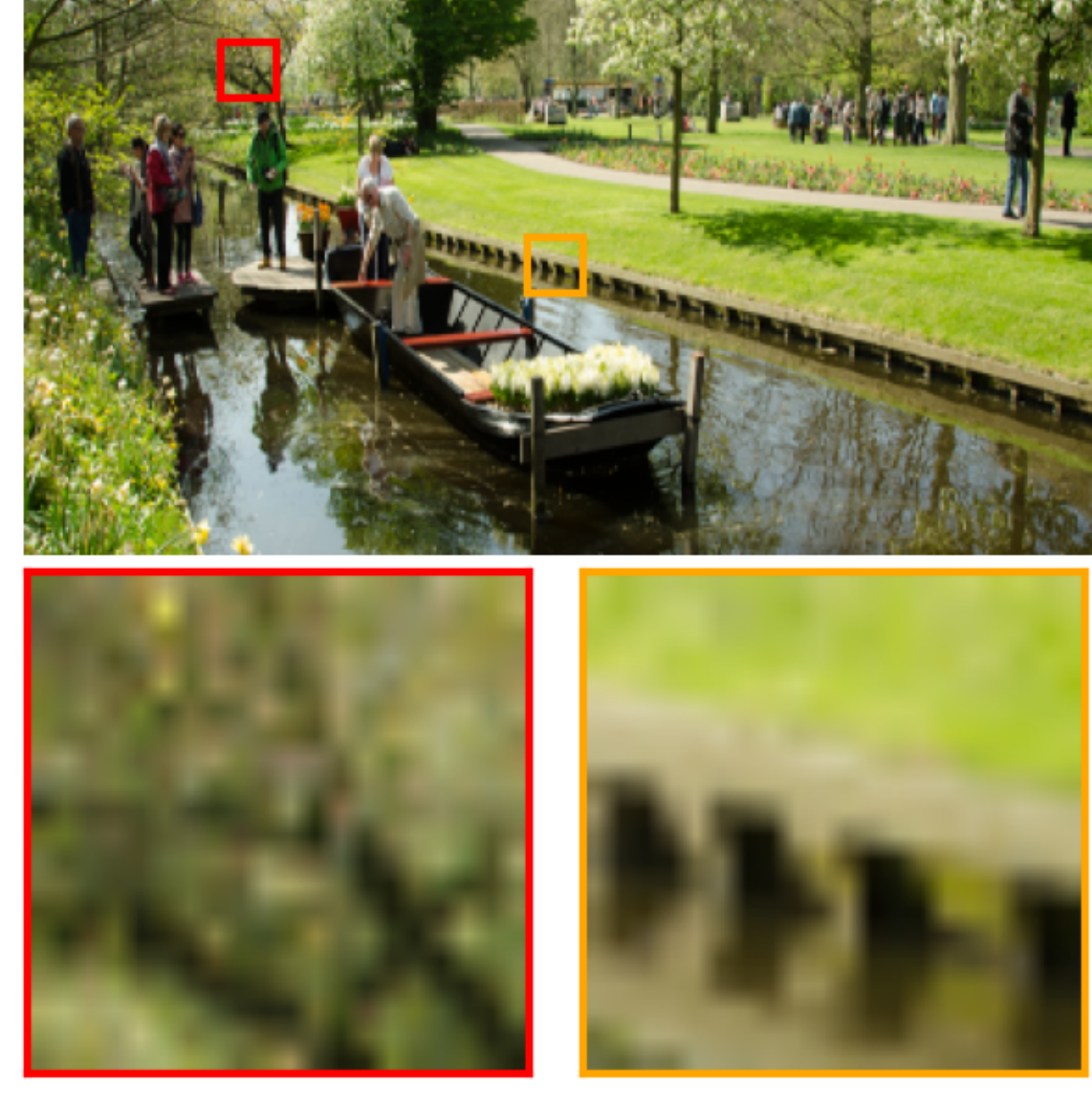}}
 	\vspace{0.2pt}
 	\centerline{\includegraphics[width=\textwidth]{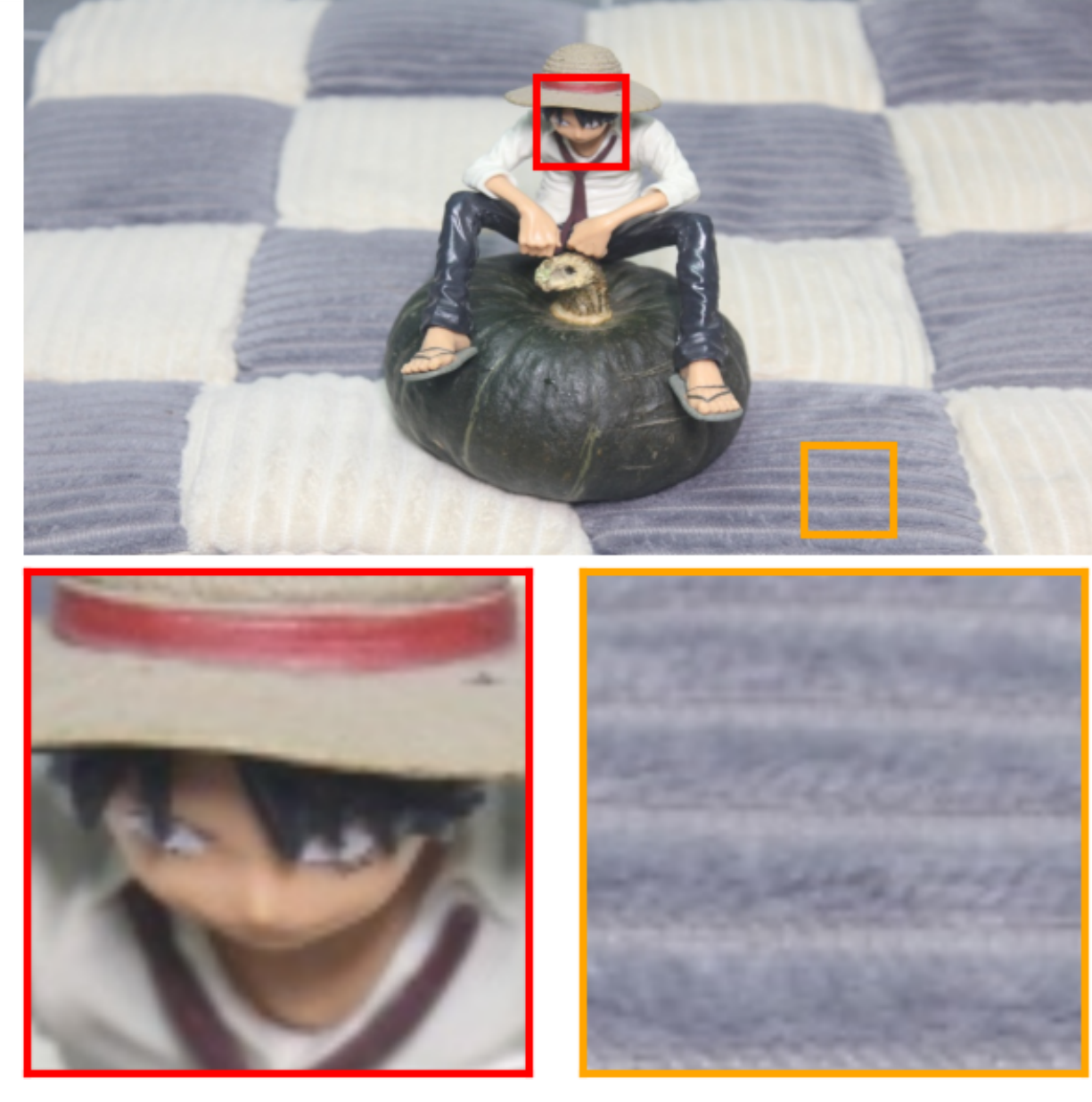}}
    \vspace{-1mm}
    \centerline{\footnotesize GroundTruth}
    \end{minipage}

    \vspace{-1.5mm}
\caption{Comparison of the enhanced images with various SOTA methods on LOL-v2-Synthetic (top row) and VILNC-Indoor (bottom row).}
    \vspace{-2mm}
\label{fig:3}
\end{figure*}

\noindent\textbf{CG-MSA.} As displayed in Fig. \ref{fig:2}(c1-c2), ${\hat {\mathbf{F}}^A} \in {{\cal R}^{C \times H \times W}}$ and ${\hat {\mathbf{F}}^B} \in {{\cal R}^{C \times H \times W}}$ are fed into the CG-MSA which is a core component for all variants of CGAB. To calculate the cross-correlation features between $\hat {\mathbf{F}}^A$ and $\hat {\mathbf{F}}^B$, we first utilize the pair downsampler \cite{13} to divide $\hat {\mathbf{F}}$ into non-overlapping 2$\times$2 patches, and obtain $\hat {\mathbf{F}}_i^j \in {{\cal R}^{C \times \frac{H}{2} \times \frac{W}{2}}}$, where $i$=1,2; $j$=$A$, $B$. Subsequently, we concatenate $\hat {\mathbf{F}}_1^A,\hat {\mathbf{F}}_2^B$ and $\hat {\mathbf{F}}_2^A,\hat {\mathbf{F}}_1^B$ along the channel dimension to generate combined features $\hat {\mathbf{F}}_i^M \in {{\cal R}^{2C \times \frac{H}{2} \times \frac{W}{2}}}$ where $i$=1, 2. Then, utilizing the SAM, the combined features are further fused to produce the cross-guided attention map. In our designed MSA, the query ($\mathbf{Q}$), key ($\mathbf{K}$) and value ($\mathbf{V}$) projections are calculated by a projective $conv1\!\times\!1$ followed by a 3$\times$3 depth-wise convolution ($dwconv3\!\times\!3$). The $conv1\!\times\!1$ is used to polymerize pixel-wise cross-channel information and then we utilize $dwconv3\!\times\!3$ to enhance the polymerized information. To avoid attention bias caused by different learning capability, the formulas for calculating the projections are as follows:
% \vspace{-1mm}
\begin{equation}
\mathbf{Q}{{\mathbf{V}}^*} = W_d^{K{V^*}}W_p^{K{V^*}}\hat {\mathbf{F}}_1^M,KV = W_d^{KV}W_p^{KV}\hat {\mathbf{F}}_2^M,
\label{eq:7}
\end{equation}
% \vspace{-1mm}
where $W_p^{(\cdot)}$ and $W_d^{(\cdot)}$ represent the learnable parameters of the $conv1\!\times\!1$ and $dwconv3\!\times\!3$, respectively. Subsequently, splitting $\mathbf{Q}{\mathbf{V}^*}$ and $\mathbf{KV}$ into $\mathbf{Q,K,V,{V}}^* \in {{\cal R}^{C \times \frac{H}{2} \times \frac{W}{2}}}$, we combine the ${\mathbf{V}^*}$ and ${\mathbf{V}}$ to produce ${\mathbf{V}^{'}}$ by using a $conv3\!\times\!3$ and divide the number of channels into each “head”. Then, the cross-guided attention map for each head can be formulated as
% \vspace{-1mm}
\begin{equation}
{\rm Attention}({\mathbf{Q}_j},{\mathbf{K}_j},{\mathbf{V}}_j^{'})\!=\!{\mathbf{V}}_j^{'}\!\odot\!{\rm Softmax}(\frac{{{\mathbf{Q}_j}\mathbf{K}_j^T}}{{{\alpha_H}}}),
\label{eq:8}
\end{equation}
% \vspace{-1mm}
where ${\mathbf{Q}}_j,{\mathbf{K}}_j,{\mathbf{V}}_j^{'}  \in {{\cal R}^{{h_k} \times \frac{{HW}}{4}}}$, ${h_k} = \frac{C}{k}$ , $j$ = 1, 2, ..., $k$; $k$ is the head number and ${\alpha _H}$ is a learnable scaling parameter that adaptively scales the magnitude of the dot product of ${\mathbf{Q}_j}$ and $\mathbf{K}_j^T$ \cite{14,15}. The $k$-head features are concatenated along the channel dimension and reshaped to $\hat {\mathbf{F}}_{out}^M \in {{\cal R}^{2C \times \frac{H}{2} \times \frac{W}{2}}}$; $\hat {\mathbf{F}}_{out}^M$ undergoes the pixel-shuffle $conv1\!\times\!1$ to upscale its resolution, and finally adds a positional encoding calculated by a $conv1\!\times\!1$ that concatenates ${\hat {\mathbf{F}}^A}$ and ${\hat {\mathbf{F}}^B}$ as input, to generate the cross-guided features $\hat {\mathbf{F}}_{out}^{AB} \in {{\cal R}^{2C \times H \times W}}$.

To simplify the analysis of computational complexity, we focus on computation of our SAM in \cref{eq:8}, which involves two matrix multiplications ${{\cal R}^{{h_k} \times \frac{{HW}}{4}}} \times {{\cal R}^{\frac{{HW}}{4} \times {h_k}}}$ and ${{\cal R}^{\frac{{HW}}{4} \times {h_k}}} \times {{\cal R}^{{h_k} \times {h_k}}}$. The formulas of the CG-MSA computational complexity are thus as follows,
% \vspace{-4mm}
\begin{equation}
\begin{split}
&{\cal O}(\mathbf{CG}{-}\mathbf{MSA})= k{\cdot}(\frac{{HW}}{4} {\cdot} {h_k} {\cdot} {h_k}) + k {\cdot} ({h_k} {\cdot} {h_k} {\cdot} \frac{{HW}}{4})\\
&= k\frac{{HW}}{2}h_k^2 = k\frac{{HW}}{2}{(\frac{C}{k})^2}= \frac{{HW{C^2}}}{{2k}}
\end{split}
\label{eq:9}
\end{equation}
% \vspace{-2mm}
Compared with the previous CNN-Transformer methods \cite{12,16}, the computational complexity of CG-MSA is linear in the spatial size, and only 25\% of the conventional MSA computation is required for the same input size. Therefore, CG-MSA can be effective in improving our TPCNet's inference speed, and can serve as a plug-and-play lightweight design module.
% \vspace{-2mm}
\section{Experiment}
% \vspace{-1mm}
\label{sec:Experiment}
\subsection{Datasets and Settings Details}
% \vspace{-1mm}

We evaluated our method on commonly-used LLIE benchmark datasets including LOL v2 \cite{29}, DICM \cite{38}, LIME \cite{11_39}, MEF \cite{40}, NPE \cite{41}, and VV \cite{42}. We also conducted further experiments on a real-world LLIE dataset VILNC \cite{25}, an extreme dataset SID (Sony-Total-Dark) \cite{43}, and an LCDP\cite{55} dataset containing over/under exposure images.

\textbf{LOL-v2.} LOL-v2 is divided into two subsets, LOL-v2-real and LOL-v2-Synthetic. The training and testing parts of LOL-v2-Real and LOL-v2-Synthetic are split proportionally into 689:100 and 900:100, respectively.

\textbf{VILNC.} VILNC is a new LLIE dataset \cite{25}, which comprises 115 indoor scenes and 20 outdoor scenes. Each indoor scene possesses three different brightness levels to simulate varying degrees of low-light images, with a normal brightness image used as a reference, while each outdoor scene only contains a pair of low/normal brightness images. We randomly extract 20\% of the images from the indoor scene as the testing set, and the rest as the training set. To compare the performance of various methods on VILNC dataset fairly, we retrain all methods in VILNC-Indoor by using their open-source code with default training parameters. Except for the method \cite{2}, QuadPrior, which is trained with large-scale datasets, we directly use its pre-training weight for inference.

\begin{table*}[t]
\captionsetup{font=small} % 单独这张表格的caption字体
\caption{Quantitative results for the LCDP \cite{55} , Sony-Total-Dark \cite{43}, MEF \cite{40}, NPE \cite{41}, LIME \cite{11_39}, DICM \cite{38}, and VV \cite{42} datasets. The top-ranking score is in \textbf{bold}, and the second-ranking is \underline{underlined}; ($\cdot$) represents the rank order.}
\vspace{-1mm}
\tiny % 或者 \footnotesize, \scriptsize, \tiny
\centering
\resizebox{\textwidth}{!}{%
\begin{tabular}{!{\vrule width 0.6pt}cc|cc|cc!{\vrule width 0.6pt}cc|ccc|c|c!{\vrule width 0.6pt}}

% \hdashline   % 顶部虚线
\noalign{\hrule height 0.6pt} % 横线粗细1pt
\multicolumn{2}{!{\vrule width 0.6pt}c|}{\multirow{2}{*}{\textbf{Methods}}}& \multicolumn{2}{c|}{\textbf{Sony-Total-Dark}} & \multicolumn{2}{c!{\vrule width 0.6pt}}{\textbf{LCDP}}& \multicolumn{2}{c|}{\multirow{2}{*}{\textbf{Methods}}} & \multicolumn{3}{c|}{\textbf{Unpaired}} &\multicolumn{1}{c|}{\multirow{2}{*}{\textbf{FLOPs(G)}}} & \multicolumn{1}{c!{\vrule width 0.6pt}}{\multirow{2}{*}{\textbf{Overall rank}}}\\
&&PSNR$\uparrow$&SSIM$\uparrow$ &PSNR$\uparrow$&SSIM$\uparrow$&&& NIQE$\downarrow$ & MUSIQ$\uparrow$ & PI$\downarrow$ & &\\
% % \hline
\specialrule{0.6pt}{1pt}{1pt} % 第一个参数：粗细, 后两个：上下间距
\multicolumn{2}{!{\vrule width 0.6pt}c|}{RUAS \cite{57}}& 10.456 & 0.086 & 13.981 & 0.633 &\multicolumn{2}{c|}{RUAS \cite{57}}& 5.248(6) & 51.275(5) & 3.955(6) & 0.83(2)&5\\
\multicolumn{2}{!{\vrule width 0.6pt}c|}{PairLIE \cite{60}}& 15.340 & 0.532 & 14.748 & 0.686 &\multicolumn{2}{c|}{ZERO-IG \cite{25}}& 3.987(4) & 47.618(6) & 3.542(5) & 11.37(4)&5\\
\multicolumn{2}{!{\vrule width 0.6pt}c|}{RetinexNet \cite{3}}& 18.230 & 0.596 & 19.626 & 0.689 &\multicolumn{2}{c|}{LANet \cite{62}}& \textbf{3.502}(1) & 56.218(3) & \textbf{2.804}(1) & 43.23(5)&\underline{2}\\
\multicolumn{2}{!{\vrule width 0.6pt}c|}{ZeroDCE++ \cite{58}}& 12.683 & 0.082 & 12.223 & 0.668 &\multicolumn{2}{c|}{ZeroDCE++ \cite{58}}& 3.959(3) & 52.172(4) & 3.231(3) & 0.61(1)&3\\
\multicolumn{2}{!{\vrule width 0.6pt}c|}{HVI-CIDNet \cite{6}}& \underline{22.904} & \underline{0.676} & \underline{22.840} & \underline{0.846} &\multicolumn{2}{c|}{QuadPrior \cite{2}}& 4.666(5) & \textbf{62.241}(1) & 3.455(4) & 1103.2(6)&4\\
\specialrule{0.6pt}{1.2pt}{0.7pt} % 第一个参数：粗细, 后两个：上下间距
\multicolumn{2}{!{\vrule width 0.6pt}c|}{\textbf{TPCNet(Ours)}}& \textbf{23.577} & \textbf{0.691} & \textbf{23.398} & \textbf{0.874}&\multicolumn{2}{c|}{\textbf{TPCNet(Ours)}}& \underline{3.625}(2) & \underline{58.070}(2) & \underline{2.982}(2) & 8.76(3)&\textbf{1} \\
\noalign{\hrule height 0.6pt} % 横线粗细1pt

\end{tabular}}
% \vspace{-10mm} % 表格下方与正文的间距
\label{table:2}
\end{table*}
\begin{figure*}[htbp]
    \vspace{-3mm}
  \centering
   \includegraphics[width=\linewidth]{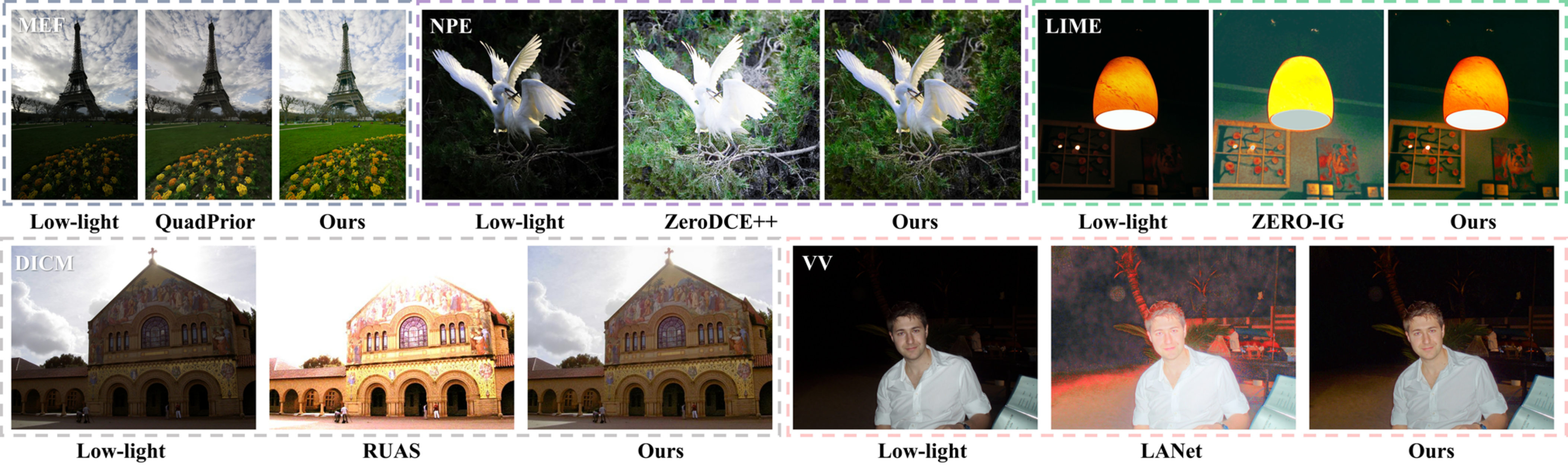}
    \vspace{-20pt}
   \caption{Visual comparison on the MEF \cite{40}, NPE \cite{41}, LIME \cite{11_39}, DICM \cite{38}, and VV \cite{42} datasets.}
   \label{fig:4}
   % \vspace{-1mm}
\end{figure*}

\textbf{SID.} The official SID dataset \cite{43} contains 5094 raw short-exposure images, each with a corresponding long-exposure reference image. We adopt its customized version subset (Sony-Total-Dark) \cite{6}, which transfers the raw format images to sRGB images without gamma correction.

\textbf{LCDP.} The LCDP \cite{55} contains images with both over-exposed and under-exposed regions, comprising 1,733 images, which are split into 1,415 for training, 100 for validation, and 218 for testing.

\textbf{Implementation Details.} Our model TPCNet is implemented by PyTorch and is trained with the Adam optimizer (${\beta _1}$= 0.9 and ${\beta _2}$ = 0.999) for 1500 epochs by using a single NIVIDA 3090 or 4090 GPU. The initial learning rate is set to $2.5\!\times\!{10^{ - 4}}$ and then steadily decreased to $1\!\times\!{10^{ - 7}}$ by the cosine annealing scheme \cite{47} during the training process. Patches of size $320\!\times\!320$ are randomly cropped from the low-/normal-light image pairs and the batch size is 8. The training data is augmented with flipping and random rotation. Inspired by reference \cite{6}, we utilize the loss function (L1loss, perceptual loss \cite{44}, edge loss \cite{46}, and SSIM loss \cite{45}) to supervise the enhanced result simultaneously in RGB space and specific color space.

\textbf{Evaluation Metrics.} For datasets with reference images, we exploit the peak signal-to-noise ratio (PSNR) and structural similarity (SSIM) \cite{48} as the pixel-level evaluation and learned perceptual image patch similarity (LPIPS) \cite{49} with AlexNet \cite{50} as the perceptual quality evaluation. For the no-reference datasets, we evaluate the generalization of models trained on LOL-v2-Synthetic using various evaluation methods, such as natural image quality evaluator (NIQE) \cite{51}, multi-scale image quality transformer (MUSIQ) \cite{52}, and perceptual index (PI) \cite{53}. We exploit PyTorch Toolbox \cite{54} for calculating no-reference metrics.
% \vspace{-3mm}
\subsection{Main Results}
% \vspace{-1.5mm}

\noindent\textbf{Quantitative Results.} The quantitative results between our TPCNet and a wide range of SOTA enhancement algorithms are shown in Tables \ref{tab:1} and \ref{table:2}. Our TPCNet outperforms SOTA methods on LOL-v2 and VILNC-Indoor datasets in terms of almost all PSNR, SSIM, and LPIPS metrics, and possesses advantages in various metrics, such as NIQE, MUSIQ, and PI on no-reference datasets (MEF \cite{40}, NPE \cite{41}, LIME \cite{11_39}, DICM \cite{38}, and VV \cite{42}), while requiring less computation. 

\begin{table*}[t]
\caption{Ablation on the LOLv2-real dataset. PSNR, SSIM, LPIPS, Params, and FLOPS (size = 256×256) are reported. The best results are marked in \textbf{bold} and the second best are \underline{underlined}.}
% \vspace{-1mm}
\centering
\resizebox{\textwidth}{!}{%
\begin{tabular}{c c}  % 三个小表格并排
\begin{tabular}{cccccccccc}
\toprule
\rowcolor{gray!20}
Baseline-1 & CAM & LFE & RFE & TPC & PSNR$\uparrow$ & SSIM$\uparrow$ & LPIPS$\downarrow$& Params (M) & FLOPs (G)\\
\rowcolor{white}
\midrule
\specialrule{0.0pt}{0.0ex}{1ex}
\checkmark &            &     &&       & 21.637 & 0.854 & 0.119 & 1.65 &5.72\\
\checkmark & \checkmark &    &&        & 23.729 & 0.871 & 0.111 & 2.61 &8.32\\
\checkmark & \checkmark & \checkmark   &&  & 24.232 & 0.876 & 0.111 & 2.61 &8.46\\
\checkmark & \checkmark & &\checkmark& & 24.263 & 0.874 & 0.109 & 2.62 &8.55\\
\checkmark & \checkmark & \checkmark &\checkmark && \underline{24.357} & \underline{0.881} & \best{0.099} & 2.62&8.68\\
\checkmark & \checkmark & \checkmark &\checkmark &\checkmark & \best{24.641} & \best{0.881} & \underline{0.100} & 2.62&8.76
\\
% \specialrule{0.0pt}{1ex}{0ex}
\specialrule{0.0pt}{1ex}{0.0ex}
\bottomrule
\bottomrule
% \specialrule{0pt}{0.35ex}{0.35ex}
\end{tabular}
&
% ---------(b)---------
\begin{tabular}{llccccc}
\toprule
\multicolumn{2}{l}{\cellcolor{gray!20}Metrics} & \cellcolor{gray!20}{PSNR$\uparrow$} & \cellcolor{gray!20}SSIM$\uparrow$ & \cellcolor{gray!20}LPIPS$\downarrow$& \cellcolor{gray!20}Params (M) & \cellcolor{gray!20}FLOPs (G)\\
\midrule
\multirow{3}{*}{\textbf{Physical Constraint}}

& w/o Eq. (3-5)            & 24.357 & 0.881 & \best{0.099}&2.62&8.68\\
&w/o Eq. (3)                            & 24.598 & 0.881 & 0.101 &2.62&8.76\\
&w/o Eq. (4-5)     & 24.575 & 0.876 & 0.110 &2.62&8.68\\
% \midrule 
\specialrule{0.1pt}{0.5ex}{0.5ex}
% \addlinespace[1ex]
\multirow{3}{*}{\textbf{Color Space}}
&HVI            & \underline{24.641} & \underline{0.881} & \underline{0.100} &2.62&8.76\\
&LAB            & 24.588 & 0.873 & 0.114 &2.62&8.76\\
&YCBCR           & \best{24.978} & \best{0.882} & 0.102 &2.62&8.76\\
% \midrule
\bottomrule
\bottomrule
% \specialrule{0pt}{0.35ex}{0.35ex}
\end{tabular}
% \vspace{-2mm}
\\
\\
\Large(a) Decomposed ablation of each component. &
\Large(b) Ablation studies on physical constraints and color space. 
\end{tabular}}
\vspace{-6mm}
\label{tab:3}
\end{table*}

For the reference datasets shown in Table \ref{tab:1}, we compared our results with the SOTA lightweight method HVI-CIDNet. Our TPCNet achieves 0.867 dB, 0.327 dB, and 0.978 dB improvements for approximately the same computational cost. Compared to RetinexFormer, a SOTA Retinex-based method, TPCNet realizes higher quality image enhancement and further expands its parameters 1.6 times (2.62/1.61), but only costs 56\% (8.76/15.57) FLOPs, demonstrating that our CG-MSA is efficiently designed. Especially, in the real-world dataset VILNC-Indoor, TPCNet achieves over 15 dB improvement compared to the previous Retinex-based methods. 

To verify TPCNet generalization on the no-reference datasets shown in Table \ref{table:2}, we compare our model with a recent SOTA unsupervised method. Compared to QuadPrior, which is trained on the large-scale datasets, TPCNet exhibits an obvious improvement in the NIQE and PI metric but only requires 0.8\% (8.76/1103.2) of its FLOPs. Compared with other approaches, TPCNet shows the advantages in NIQE, MUSIQ, and PI metrics without a large computational consumption, achieving a balance between performance and efficiency. These results demonstrate that the TPC constructed on an implicit feature space can improve the generalization and robustness of our model.

\noindent\textbf{Qualitative Results.} The visual results for comparing TPCNet to other SOTA algorithms are shown in Figs. \ref{fig:3} and \ref{fig:4}. From the Fig. \ref{fig:3} visualization, we can observe that TPCNet achieves stable brightness enhancement and maintains color consistency while other algorithms exhibit a certain flaw, especially in real-world datasets (VILNC-Indoor). For example, the earliest Retinex-based DL, RetinexNet, produces serious color deviation during enhancement when facing extreme darkness, while RetinexFormer recovers the correct color, but still with a little color deviation. Compared to a recent SOTA algorithm, HVI-CIDNet can recover accurate color due to its HVI color space advantage, but some details are blurred.

In addition, visual comparisons between TPCNet and other SOTA supervised and unsupervised algorithms on five datasets without ground truth are illustrated in Fig. \ref{fig:4}. For each dataset, we select one image to compare our method with a SOTA algorithm. From the results, we can see the shortcomings of the previous SOTA methods during the enhancement process, for example, overexposure/underexposure, color deviation, and amplified noise. In contrast, our TPCNet can effectively enhance low-light regions without introducing obvious spots and serious color deviation. These quantitative and qualitative results demonstrate the superior performance of TPCNet over previous SOTA methods in the lightweight LLIE field and demonstrate that the TPC constructed in implicit space can improve the performance and generalization of our model.
% \vspace{-3mm}
\subsection{Ablation Study}
% \vspace{-1mm}

We perform all ablation experiments on LOL-v2-Real for good convergence and stable performance of TPCNet. To verify the key modules, physical constraint, and the CAM with different color spaces in TPCNet, we use quantitative comparisons. The results are reported in Table \ref{tab:3}. 

\noindent\textbf{Decomposed ablation.} We utilize a decomposed ablation to analyze the effect of each component on improving performance, as displayed in Table \ref{tab:3}a. To obtain Baseline-1, which only comprises the DCGT framework, we remove the CAM from TPCNet and replace its LFE and RFE with $conv3\!\times\!3$. Meanwhile, \cref{eq:3} is replaced by a $conv3\!\times\!3$ mapping the output from the last CGAB to the enhanced result, and \cref{eq:4,eq:5} are removed in TPCNet, to avoid the impact of physical constraint. When we exploit the CAM with HVI \cite{6} as the color space, the mapping $conv3\!\times\!3$ is utilized to produce enhanced brightness instead of the final enhanced result. Baseline-1 realizes an obvious improvement, gaining 2.102 dB after applying CAM. Subsequently, we employ the LFE and RFE, respectively, and Baseline-1 achieves 2.605 and 2.636 dB improvements, respectively. When jointly applying these two estimators, the Baseline-1 achieves an increase of 2.730 dB. Subsequently, the TPC is introduced to obtain our final TPCNet version, and Baseline-1 realizes a 3.004 dB improvement. These decomposed experiments prove the effectiveness of CAM, LFE, RFE, and physical constraint in our TPCNet.
% \begin{figure}[htb]
%   \centering
%   % \fbox{\rule{0pt}{2in} \rule{0.9\linewidth}{0pt}}
%    \includegraphics[width=\linewidth]{Ablation_图片2.jpg}
%     \vspace{-20pt}
%    \caption{aa}
%    \label{fig:5}
% \end{figure}
% \begin{figure}[htb]
%   \centering
%   % \fbox{\rule{0pt}{2in} \rule{0.9\linewidth}{0pt}}
%    \includegraphics[width=\linewidth]{Ablation_图片3.jpg}
%     \vspace{-20pt}
%    \caption{aa}
%    \label{fig:6}
% \end{figure}

\noindent\textbf{Physical constraint.} To analyze the effect of TPC, we administer an ablation experiment to study each physical constraint. The quantitative results are exhibited in Table \ref{tab:3}b We first remove TPC from TPCNet, and initially obtain an PSNR of 24.357 dB. Then, we separately apply \cref{eq:4,eq:5} and \cref{eq:3} to form constraints on the inner network, and the model achieves a gain of 0.241 dB and 0.218 dB, respectively. Finally, we construct the complete TPC for further improvement. These results validate that exploiting physical constraints within the inner network can optimize performance without requiring additional parameters, simultaneously providing an interpretable method for network design. 

\noindent\textbf{Color spaces.} We conduct ablation experiments to study the robustness of CAM with different color spaces and further explore its impact on model performance. Based on our image formation model, we chose a color conversion that can transform RGB images into a color space where brightness and color information can be separated, such as HVI \cite{6}, YCbCr \cite{34}, and LAB \cite{35}. The corresponding quantitative results are presented in Table. \ref{tab:1}b. Our results demonstrate the great robustness of CAM in various color spaces, and can achieve superior performance (the PSNR is 24.641 / 24.588 / 24.978 with Color space HVI / LAB / Ycbcr), better than the SOTA methods \cite{6,12,16} in recent years.
% \vspace{-4mm}
\section{Conclusion}
% \vspace{-1mm}
\label{sec:Conclusion}
In this work, we have reformulated the influence of physical constraints in the imaging process based on the Kubelka-Munk theory, and have constructed a the physical constraint among illumination, reflection, and detection. Our TPC theory addresses the limitations of previous Retinex-based algorithms, which view the reflected objects as ideal Lambertian and ignore the specular reflection coefficient in the modeling process. Extensive quantitative and qualitative experiments verify the effectiveness of our model in improving the robustness and LLIE ability by constructing physical constraints in feature space, and demonstrate that TPCNet outperforms SOTA LLIE methods on 10 datasets, thus expanding the interpretable network-design direction for low-light enhancement.